\documentclass{article}

\usepackage[preprint]{neurips_2024}

\usepackage[utf8]{inputenc} 
\usepackage[T1]{fontenc}    
\usepackage{hyperref}       
\usepackage{url}            
\usepackage{booktabs}       
\usepackage{amsfonts}       
\usepackage{nicefrac}       
\usepackage{microtype}      
\usepackage{xcolor}         
\usepackage{graphicx}
\usepackage{amsmath}
\usepackage{amssymb}
\usepackage{amsthm}
\usepackage{multirow}
\usepackage{subfigure}
\usepackage{algorithmicx}
\usepackage[ruled,vlined]{algorithm2e}
\title{Multi-label Class Incremental Emotion Decoding with Augmented Emotional Semantics Learning}

\author{%
	Kaicheng Fu $^{1, 2}$\thanks{Equal contributions}, \enspace Changde Du$^{1, 2}$\footnotemark[1], \enspace Xiaoyu Chen$^{1,2}$, \enspace Jie Peng$^{1,2}$, \enspace
	Huiguang He$^{1, 2}$\thanks{Huiguang He is the corresponding author.}\\
	$^1$Key Laboratory of Brain Cognition and Brain-inspired Intelligence Technology,\\
	Institute of Automation, Chinese Academy of Sciences\\
	$^2$School of Artificial Intelligence, University of Chinese Academy of Sciences\\
	\texttt{fukaicheng2019,changde.du,chenxiaoyu2022,pengjie2023,huiguang.he@ia.ac.cn}
}

\begin{document}

\maketitle

\begin{abstract}
Emotion decoding plays an important role in affective human-computer interaction. However, previous studies ignored the dynamic real-world scenario, where human experience a blend of multiple emotions which are incrementally integrated into the model, leading to the multi-label class incremental learning (MLCIL) problem. Existing methods have difficulty in solving MLCIL issue due to notorious catastrophic forgetting caused by partial label problem and inadequate label semantics mining. In this paper, we propose an augmented emotional semantics learning framework for multi-label class incremental emotion decoding. 
Specifically, we design an augmented emotional relation graph module with label disambiguation to handle the past-missing partial label problem.
Then, we leverage domain knowledge from affective dimension space to alleviate future-missing partial label problem by knowledge distillation.
Besides, an emotional semantics learning module is constructed with a graph autoencoder to obtain emotion embeddings in order to guide the semantic-specific feature decoupling for better multi-label learning.
Extensive experiments on three datasets show the superiority of our method for improving emotion decoding performance and mitigating forgetting on MLCIL problem.
\end{abstract}

\section{Introduction}
The primary challenge in affective \textbf{h}uman-\textbf{c}omputer \textbf{i}nteraction (HCI) research lies in accurately decoding the human emotional states. Although conventional deep learning based solutions have achieved high performance \cite{li2020deep,poria2017review}, these methods fail to adapt sufficiently to the dynamics of real-world scenario. On one hand, with the advancement of psychology, new emotion categories are continuously discovered to depict more fine-grained emotional experience. On the other hand, affective HCI robots need to provide personalized emotional support to different demographics. Thus, it is imperative for affective HCI systems to incrementally integrate novel information, such as new emotion categories.


Furthermore, a famous novelist Jeffrey Eugenides \cite{eugenides2003middlesex} once said ``\emph{Emotions, in my experience, are not covered by single words. I do not believe in 'sadness', 'joy', or 'regret'}", which illustrates that human emotions are often complicated hybrid. People usually perceive a blend of multiple emotions when receiving emotional stimulation in their daily life, which brings about multi-label emotion decoding problem \cite{fu2022multi}. Based on this, we put forward a novel research problem named \emph{multi-label fine-grained class incremental emotion decoding}.

\begin{figure*}[t] 
	\centering
	\includegraphics[width=0.7\linewidth]{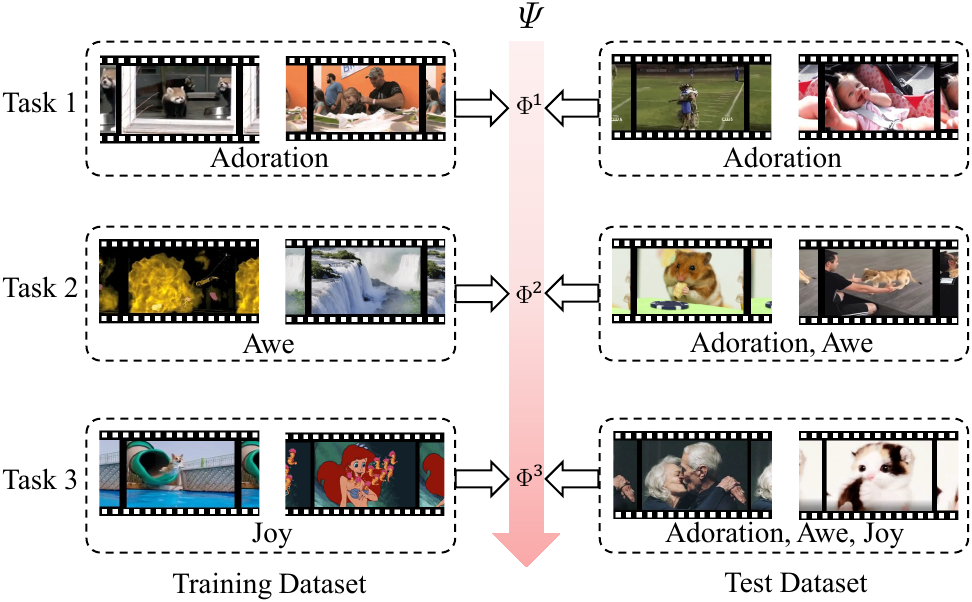}
	\caption{The illustration of the multi-label class-incremental emotion decoding task. Supposed there are three emotion categories in total: \emph{adoration}, \emph{awe}, and \emph{joy}, which are incrementally learned in three tasks. $\Phi^1$,$\Phi^2$,$\Phi^3$ are the parameters of the unified model $\psi$, which are continuously trained.}
	\label{fig:mi}
\end{figure*}

As shown in Figure \ref{fig:mi}, multi-label class incremental emotion decoding aims to continually construct a unified model that can incrementally learn and integrate the knowledge of old and new classes and make a comprehensive decoding of multiple emotion categories for an instance. Compared with \textbf{s}ingle-\textbf{l}abel \textbf{c}lass \textbf{i}ncremental \textbf{l}earning (SLCIL), the catastrophic forgetting issue faced by \textbf{m}ulti-\textbf{l}abel \textbf{c}lass \textbf{i}ncremental \textbf{l}earning (MLCIL) mainly arises from the challenge of past- and future-missing partial label problem. 
Regarding the past-missing partial label problem, the left screenshot (\emph{The corgi is diving}) in the training dataset for task $3$ also has the label \emph{Adoration}. Nevertheless, this label is unseen for model during current task. 
As for the future-missing partial label problem, the right screenshot (\emph{The father is combing his daughter's hair}) in the training dataset for task $1$ also has the label \emph{Joy}. However, model can not have access to this label in current task. 
Previous studies for MLCIL have either stored some old instances \cite{kim2020imbalanced,liang2022optimizing}, which limits its practical application, or neglected the issue of future-missing partial label problem \cite{du2022agcn} resulting in suboptimal results.

To tackle the above challenges, a novel \textbf{a}ugmented \textbf{e}motional \textbf{s}emantics \textbf{l}eanring (AESL) framework is proposed.
Firstly, for the past-missing partial label problem, we design an augmented \textbf{e}motional \textbf{r}elation \textbf{g}raph (ERG) module with graph-based label disambiguation. When new task is coming, this module not only generates reliable soft labels for the old emotion classes but also constructs new ERG by combining old ERG and new data to maintain crucial historical emotional label correlation information.
Secondly, for the future-missing partial label problem, affective dimension space, which is another emotion model and able to represent infinitely many emotion categories \cite{russell1977evidence}, can provide supplementary domain knowledge \cite{le2023uncertainty} for continually learning emotion categories. Inspired by this, we develop a relation-based knowledge distillation framework for aligning the model feature and affective dimension space. 
Besides, we leverage the ERG and design an emotional semantics learning module with a graph autoencoder to learn emotion embeddings for further feature decoupling, which can obtain the indispensable semantic-specific features for better multi-label learning.
For extensive evalutation, we conduct our experiments on a human brain activity dataset with 5 subjects named \emph{Brain27} and two multimedia datasets called \emph{Video27} and \emph{Audio28} on multiple incremental learning settings, which involve up to 28 fine-grained emotion categories. To summarize, our main contributions can be summarized as:
\begin{itemize}
	\item For the first time, we introduce the problem of multi-label class incremental emotion decoding, aiming to advance the research on dynamic HCI systems in real-world scenario.
	\item We develop a novel augmented emotional semantics learning framework to improve emotion decoding performance and mitigate forgetting on MLCIL problem.
	\item Sufficient experiments on three datasets and multiple incremental learning protocols  demonstrate the effectiveness of AESL. Our source code will be shared online post publication.
\end{itemize}

\begin{figure*}[t]
	\begin{center}
		\includegraphics[scale=0.4]{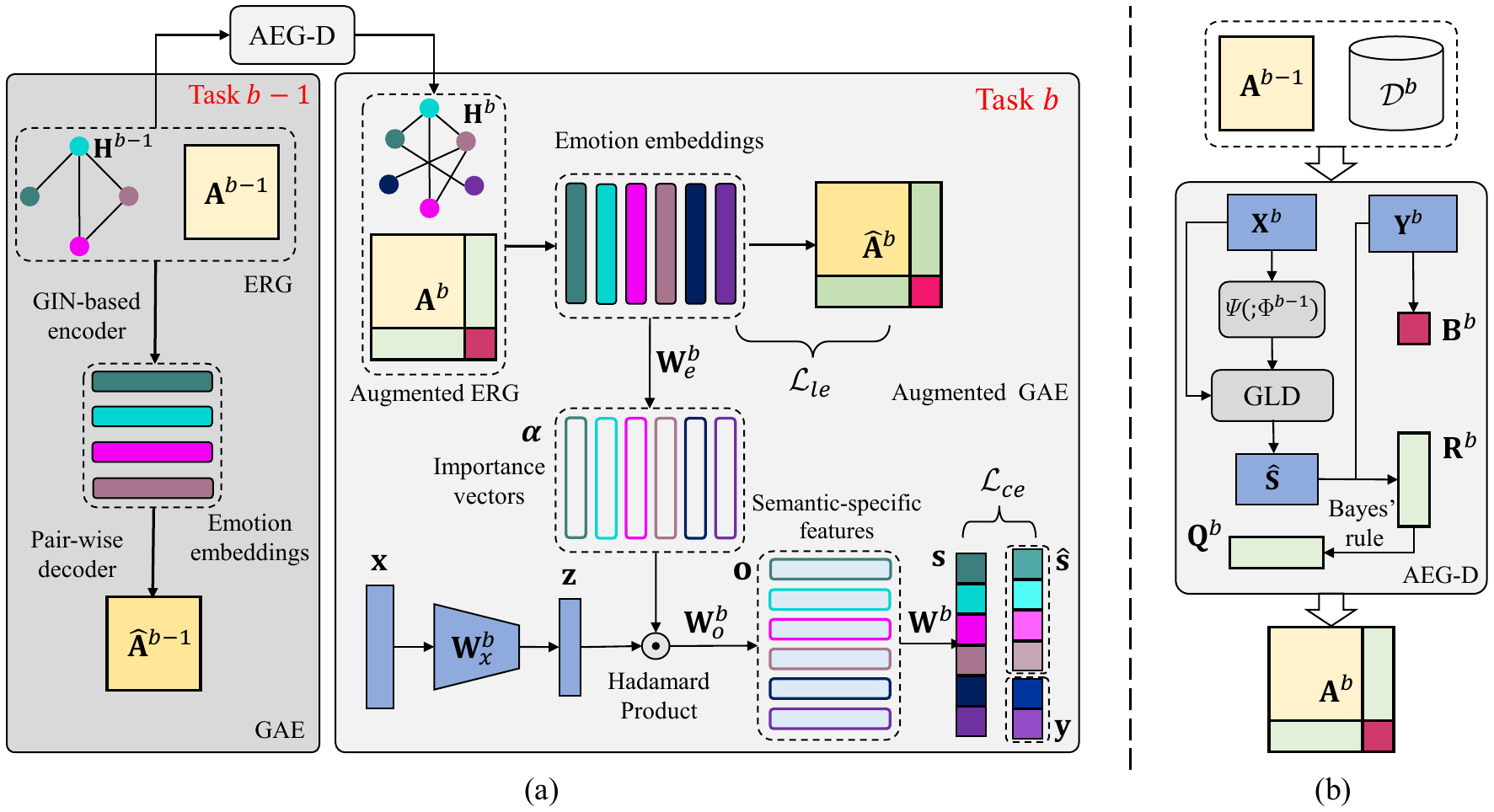}
	\end{center}
	\vspace{-0.48cm}
	\caption{The framework of AESL for multi-label class incremental emotion decoding. (a) Emotional semantics learning and semantic-guided feature decoupling procedure in incremental learning scenario. We omit semantic-guided feature decoupling module in task $b-1$ for clarity. (b) The process of constructing augmented emotional relation graph with label disambiguation in task $b$.}
	\label{fig:framework}
	\vspace{-0.48cm}
\end{figure*}

\section{Methodology}
Figure \ref{fig:framework} shows the framework of AESL. In this section, we first formalize the MLCIL problem. Then, we introduce the emotional semantics learning module and semantic-guided feature decoupling. And we show how to augment ERG during incremental learning, followed by the relation-based knowledge distillation from affective space. Finally, the overall objective function is exhibited. The detailed of AESL algorithm is written in Appendix \ref{algorithm}.
\vspace{-0.2cm}
\subsection{Problem Formulation}
In the MLCIL scenario, we assume that there is a sequence of $B$ training tasks $\left\{\mathcal{D}^{1}, \mathcal{D}^{2}, \cdots, \mathcal{D}^{B}\right\}$ without overlapping emotion classes, where $\mathcal{D}^{b}=\left\{\left(\mathbf{x}_{i}^{b}, Y_{i}^{b}\right)\right\}_{i=1}^{n^b}$ is the $b$-th incremental task with $n^b$ training instances. $\mathbf{x}_i^b \in \mathbb{R}^D$ is an instance with classes $Y_i^b \subseteq C^b$. $C^b$ is the label set of task $b$, where
$C^b  \cap C^{b^\prime} = \varnothing$ for $b\neq b^\prime$. We can only access data from $\mathcal{D}^b$ during the training of task $b$. $\mathbf{y}_{i}^{b} \in \mathbb{R}^{|C^b|}$ is the multi-hot label vector in which $y_{ic}^{b} \in \left\lbrace 0,1 \right\rbrace$. $y_{ic}^{b}=1$ means that emotion label $c$ is relevant to the instance $\mathbf{x}_i^b$. The unified model should not only acquire the knowledge from the current task $\mathcal{D}^b$ but also preserve the knowledge from former tasks. After task $b$, the trained model is evaluated over all seen emotion classes $\mathcal{C}^b=C^1 \cup \cdots C^b$.
\vspace{-0.2cm}
\subsection{Emotional Semantics Learning}
\label{lsl}
In task $b$, we firstly construct an augmented ERG, denoted as $\mathcal{G}^b = (V^b,E^b)$, where $V^b$ denotes the set of nodes corresponding to the set of class labels $\mathcal{C}^b$ and $E^b$ refers to the set of edges. The adjacency matrix $\mathbf{A}^b \in \mathbb{R}^{|\mathcal{C}^b| \times |\mathcal{C}^b|}$ stores the weights associated with each edge, indicating the degree of co-occurrence relationship between pairs of emotion labels, thus reflecting the strength of their connection. In this section, we focus on how to obtain the emotion embeddings in  task $b$ and the details of computing $\mathbf{A}^b$ will be discussed in Section \ref{aERG}.

After that, we adopt a graph autoencoder to project emotion labels to a label co-occurrence semantic space with the ERG. We exploit Graph Isomorphism Network\cite{xu2018powerful} (GIN) as the encoder of our graph autoencoder due to its powerful representation learning capability. Specifically, given a feature matrix of nodes $\mathbf{H}^b_{l} \in \mathbb{R}^{|\mathcal{C}^b|\times d_l}$ in which each row refers to the embedding of an emotion label and $d_l$ corresponds to the dimensionality of node features in $l$-th GIN layer. The node features are able to update within a GIN layer with message passing strategy by:
\begin{equation}
	\mathbf{H}^b_{l+1} = f_{l+1}[(1+\epsilon_{l+1})\mathbf{H}^b_{l} + \mathbf{A}^b\mathbf{H}^b_{l};\theta^b_{l+1}],
	\label{mp}
\end{equation}
in which $\mathbf{H}^b_{l+1}\in \mathbb{R}^{|\mathcal{C}^b|\times d_{l+1}}$ is the updated feature matrix of nodes, $f_{l+1}(:,\theta_{l+1})$ refers to a fully-connected neural network. Additionally, $\epsilon_{l+1}$ is a learnable parameter which regulates the importance of node's own features during the process of neighborhood aggregation. Unlike object labels in image classification \cite{chen2019multi}, emotion category labels are difficult to obtain initial word embeddings directly from language models. Consequently, the initial feature matrix of nodes $\mathbf{H}^b_{0}\in \mathbb{R}^{|\mathcal{C}^b|\times d_0}$ is initialized by standard Gaussian distribution and we set $d_0=|\mathcal{C}^b|$ (can also be other task-agnostic constants). After stacking $L$ GIN layers, we take use of $\mathbf{E}^b = \mathbf{H}^b_{L}  \in \mathbb{R}^{|\mathcal{C}^b|\times d_L}$ as the final emotion label semantic embeddings in task $b$ for further semantic-specific feature extraction.

Furthermore, we introduce a pairwise decoder to reconstruct the adjacency matrix $\mathbf{A}^b$, which can ensure the obtained label embeddings capture the topological structure of the label semantic space well. The loss function of the pairwise decoder can be written as:
\begin{equation}
	\mathcal{L}_{le} = \frac{1}{|\mathcal{C}^b|^2}\sum_{i=1}^{|\mathcal{C}^b|}\sum_{j=1}^{|\mathcal{C}^b|}[ \frac{(\mathbf{e}_i^b-\bar{\mathbf{e}}^b)^T(\mathbf{e}_j^b-\bar{\mathbf{e}}^b)}{||\mathbf{e}_i^b-\bar{\mathbf{e}}^b||||\mathbf{e}_j^b-\bar{\mathbf{e}}^b||}
	-\hat{\mathbf{A}}^b_{ij}]^2,
	\label{le_loss}
\end{equation}
where $\bar{\mathbf{e}}^b=\mathbb{E}_i[\mathbf{e}_i^b]$ corresponds to the average emotion embeddings in task $b$, $\mathbf{e}_i^b$ denotes the $i$-th row of $\mathbf{E}^b$, and $\hat{\mathbf{A}}^b = \mathbf{A}^b + \mathbf{I}^b$ and $\mathbf{I}^b$ is an identity matrix.

Overall, emotional semantics learning aims to fit the function $f(:,\theta^b)$, resulting in $\mathbf{E}^b=f(\mathbf{H}^b_{0},\mathbf{A}^b;\theta^b)$ in task $b$.
\subsection{Semantic-Guided Feature Decoupling}
\label{sgfd}
A multi-label classification model with semantic-guided feature decoupling can be regarded as the composition of a semantic-specific feature extractor $g$ and a classification head (omit bias for simplicity) $\mathbf{W}$, where $g(;\phi^b):\mathbb{R}^{D} \times \mathbb{R}^{|\mathcal{C}^b| \times d_L} \rightarrow \mathbb{R}^{|\mathcal{C}^b| \times d}$ and $\mathbf{W}^b\in\mathbb{R}^{d\times |\mathcal{C}^{b}|}$ in task $b$. For a specific emotion label $\ell_k \in \mathcal{C}^b$, semantic-specific mapping $g_k$ can be formulated as $g_k(\mathbf{x}^b,\mathbf{e}_k^b;\phi^b) \in \mathbb{R}^d$. The linear layer can be further decomposed in the combination of classifiers $\mathbf{W}^b=[\mathbf{w}_1,\cdots,\mathbf{w}_{|\mathcal{C}^{b}|}]$, in which each classifier corresponds to one emotion embedding. The classification head will be expanded for new classes as the continual learning progresses. The key issue is to design the semantic-specific feature extractor $g_k$.

To achieve this, we firstly map instance representation $\mathbf{x}^b$ from the original feature space to a more powerful deep latent feature $\mathbf{z}$ with a fully-connected network with parameters $\mathbf{W}_x^b \in \mathbb{R}^{D\times d_z}$ and $\mathbf{b}_x^b\in \mathbb{R}^{d_z}$. In order to utilize emotional semantics to guide the feature extraction for each label, we adopt an attention-like mechanism. Concretely, we attempt to obtain feature importance value $\boldsymbol{\alpha}_k$ for each emotion category by using a fully-connected network followed by a $\mathrm{sigmoid}$ function for $\mathbf{e}_k^b$ with parameters $\mathbf{W}_e^b \in \mathbb{R}^{d_L\times d_z}$ and $\mathbf{b}_e^b\in \mathbb{R}^{d_z}$.
Then, we select pertinent features for each emotion category via Hadamard product between the feature important vector and the latent representation. Successively, we can obtain the semantic-specific feature for each emotion category from another fully-connected network. This procedure can be formulated as follows:
\begin{equation}
	\mathbf{o}_k	= \zeta[{\mathbf{W}_o^b}^{T}(\mathbf{z}\odot \boldsymbol{\alpha}_k)+\mathbf{b}_o^b],
	\label{lsf}
\end{equation}
in which $\mathbf{W}_o^b \in \mathbb{R}^{d_z\times d}$ and $\mathbf{b}_o^b \in \mathbb{R}^{d}$ are shared learnable parameters. $\odot$ refers to the Hadamard product, and $\zeta$ denotes activation function. At this point, we have defined the semantic-specific feature extractor $\mathbf{o}_k = g_k(\mathbf{x}^b,\mathbf{e}_k^b;\phi^b)$, in which $\phi=\left\lbrace \mathbf{W}_x, \mathbf{W}_e, \mathbf{W}_o, \mathbf{b}_x, \mathbf{b}_e, \mathbf{b}_o \right\rbrace $. Then, we can predict the confidence score of the presence of emotion label $\ell_k$ through the corresponding classifier:
\begin{equation}
	\label{confidence}
	s_k = \sigma(\mathbf{w}_k^T\mathbf{o}_k+b_k) = \sigma(\mathbf{w}_k^Tg_k(\mathbf{x}^b,\mathbf{e}_k^b;\phi^b)+b_k),
\end{equation}
in which $k \in \left\lbrace 1,\cdots, |\mathcal{C}^b|\right\rbrace $.
\subsection{Augmented Emotional Relation Graph}
\label{aERG}
So far, for a given task $b$, we have implemented emotional semantics learning and utilized emotion embeddings to guide the extraction of semantic-specific features. For clarity, we denote the above procedure as $\mathbf{s} = \psi(\mathbf{x},\mathbf{A};\Phi)$, in which $\Phi = \left\lbrace \theta,\phi,\mathbf{W} \right\rbrace $. In this section, we show the \textbf{a}ugmented \textbf{e}motional relation \textbf{g}raph module with label \textbf{d}isambiguation (AEG-D) shown in Figure \ref{fig:framework} (b).

At the beginning of task $b$, we have access to new labels $C_{b}$. Compared with existing emotional relation graph $\mathcal{G}^{b-1}$, we need to augment the node set and adjacency matrix to $V^{b}$ and $\mathbf{A}^{b}$, respectively. For the former, we only need to sample $|\mathcal{C}_{b}|$ vectors from the standard Gaussian distribution.
For the latter, it is difficult to infer $\mathbf{A}^{b}$ directly from statistical label co-occurrence due to the partial label problem.

In our experiment, the adjacency matrix on the given class set $\mathcal{C}$ is defined based on label co-occurance:

\begin{equation}
	\mathbf{A}_{ij} = P(\ell_i\in \mathcal{C}|\ell_j\in \mathcal{C})|_{i\neq j} = \frac{N_{ij}}{N_{j}},
	\label{label_co}
\end{equation}
in which $N_{ij}$ is the number of instances with both class $\ell_i$ and $\ell_j$, $N_{j}$ is the number of instances with class $\ell_j$. When the task $b$ is coming, the augmented adjacency matrix $\mathbf{A}^{b}$ can be formulated as the following block form\cite{du2022agcn}:
\begin{equation}
	{\mathbf{A}}^{b}=\begin{bmatrix} {\mathbf{A}}^{b-1} & \mathbf{R}^{b} \\ \mathbf{Q}^{b} & \mathbf{B}^{b} \end{bmatrix} \Leftrightarrow \begin{bmatrix} \text{Old-Old} & \text{Old-New} \\ \text{New-Old} & \text{New-New} \end{bmatrix}.
\end{equation}
In the four blocks of matrix ${\mathbf{A}}^{b}$, $\mathbf{A}^{b-1}$ can be directly inherited from task $b$, and $\mathbf{B}^{b}$ can be easily computed from $\mathcal{D}^{b}$. However, $\mathbf{R}^{b}$ and $\mathbf{Q}^{b}$ involve the inter-task label relationship between old classes in past tasks and new classes in task $b$. We should first assign soft labels in the past label set $\mathcal{C}^{b-1}$ for the instances in new dataset $\mathcal{D}^{b}$ for subsequent calculation. Although $\mathbf{s}^{b} = \psi(\mathbf{x}^{b},\mathbf{A}^{b-1};\Phi^{b-1})$ is a feasible solution for the soft labels construction, this kind of soft labels contain a significant amount of noise and fail to utilize the correlation among instances.

To tackle this problem, we adopt a \textbf{g}raph-based \textbf{l}abel \textbf{d}isambiguation (GLD) module to the label confidence score $\mathbf{s}^b$. Firstly, the similarity between two instances is calculated with Gaussian kernel (omit $b$ without ambiguity) $\mathbf{P}_{ij} = \mathrm{exp}(-\frac{||\mathbf{x}_i-\mathbf{x}_j||^2}{2\sigma^2})$, in which $\mathbf{x}_i$ and $\mathbf{x}_j$ are two different samples in $\mathcal{D}^{b}$. Following the label propagation procedure, let $\hat{\mathbf{P}} = \mathbf{P}\mathbf{D}^{-1}$ be the propagation matrix by normalizing weight matrix $\mathbf{P}$ in column, where $\mathbf{D} = \mathrm{diag}[d_1,\cdots,d_{n^{b}}]$ is the diagonal matrix with $d_j = \sum_{i=1}^{n}\mathbf{P}_{ij}$. Assume that we have access to a past label confidence matrix using $\psi(;\Phi^{b-1})$ for $\mathcal{D}^{b}$, which denotes as $\mathbf{S} \in \mathbb{R}^{n^{b}\times|\mathcal{C}^{b-1}|}$. And we set the initial label confidence matrix $\mathbf{F}_{0}=\mathbf{S}$. For the $t$-th iteration, the refined label confidence matrix is updated by propagating current labeling confidence over $\hat{\mathbf{P}}$:
\begin{equation}
	\mathbf{F}_{t} = \beta  \cdot \hat{\mathbf{P}}^T\mathbf{F}_{t-1} + (1-\beta)  \cdot \mathbf{F}_{0}.
	\label{pro}
\end{equation}
The balancing parameter $\beta \in [0,1]$ controls the labeling information inherited from iterative label propagation and $\mathbf{F}^{0}$. Let $\mathbf{F}^{*}$ be the final label confidence matrix and also serve as the soft labels after disambiguation, which means $\hat{\mathbf{S}}=\mathbf{F}^{*}$.  We set the balancing parameter $\beta$ to 0.95 during label disambiguation according to \cite{chen2020multi}.

With the dataset $\mathcal{D}^{b}$ and soft label matrix $\hat{\mathbf{S}}$, we are able to compute $\mathbf{R}^{b}\in \mathbb{R}^{|\mathcal{C}^{b-1}|\times|C^{b}|}$ as follows:
\begin{equation}
	\mathbf{R}_{ij}^{b} = P(\ell_i\in \mathcal{C}^{b-1}|\ell_j\in C^{b}) = \frac{\sum_{\mathbf{x}}\hat{s}_iy_j}{N_{j}},
	\label{rb}
\end{equation}
in which $\hat{s}_i$ denotes the value of class $i$ corresponding to instance $\mathbf{x}$ in soft label matrix, and $y_j$ refers to the value of class $j$ corresponding to the same instance in the label matrix which derives from $\mathcal{D}^{b}$. Naturally, following the Bayes' rule, we can obtain the $\mathbf{Q}^{b}\in \mathbb{R}^{|C^{b}|\times|\mathcal{C}^{b-1}|}$ by:
\begin{equation}
    \mathbf{Q}_{ji}^{b} = P(\ell_j\in C^{b}|\ell_i\in \mathcal{C}^{b-1})= \frac{P(\ell_i\in \mathcal{C}^{b-1}|\ell_j\in C^{b})P(\ell_j\in C^{b})}{P(\ell_i\in \mathcal{C}^{b-1})}=\frac{\mathbf{R}_{ij}^{b}N_j}{\sum_{\mathbf{x}}\hat{s}_i}.
	\label{qb}
\end{equation}

Above all, we have constructed the adjacency matrix $\mathbf{A}^{b}$ and achieved continual learning new emotion categories in the multi-label scenario. It is noticeable that, in our experiments, we actually utilize symmetric adjacency matrix for model training by the $\frac{\mathbf{A}+\mathbf{A}^T}{2}$ operation.

\subsection{Relation-based Knowledge Distillation}
\label{rkd}
Although we have achieved the semantic-specific feature learning and overcome past-missing partial label problem by ALG-D, we have not yet addressed the issue of future-missing partial label problem in MLCIL. Previous studies \cite{schlosberg1954three,russell1977evidence} has shown that affective dimension, as a complementary emotion model to emotion category, can represent infinitely many emotion categories within its constructed affective space. We propose that incorporating the domain knowledge of affective dimension space into the model for alleviating the problem of future-missing partial label problem. 

\begin{figure}[t]
	\begin{center}
		\includegraphics[scale=0.5]{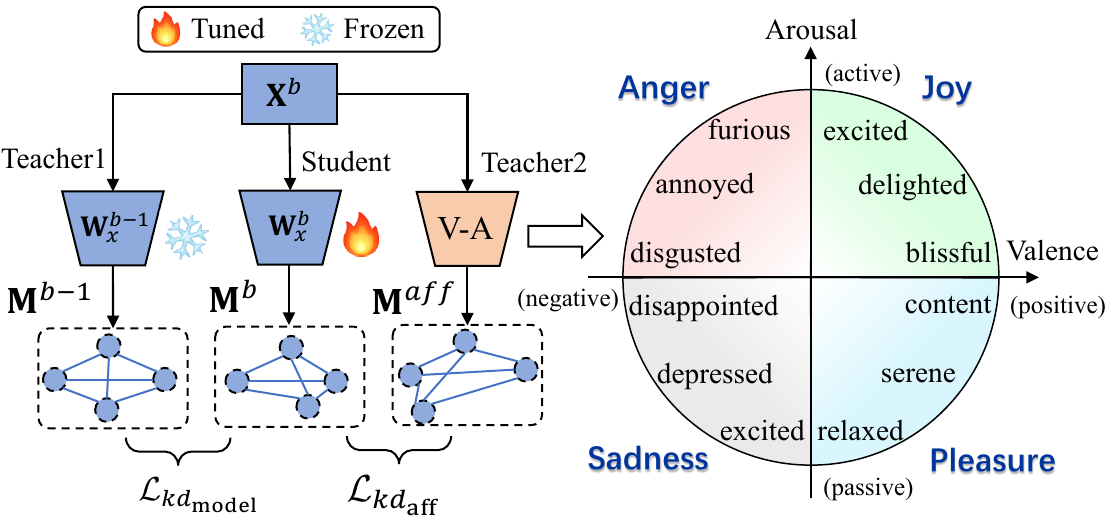}
	\end{center}
	\caption{Diagram of relation-based knowledge distillation with two teachers in the process of training task $b$. Each discrete emotion category represents a point in the affective space formed by Arousal and Valence.}
	\label{fig:kd}
\end{figure}

In terms of methodology, during the training process for each task, we attempt to align the feature space of our model with the predefined affective space constructed by some affective dimensions such as \emph{Arousal} and \emph{Valence}. Specifically, taking into account the heterogeneity of the two spaces, we adopt \textbf{r}elation-based \textbf{k}nowledge \textbf{d}istillation (RKD). We firstly calculate the \textbf{r}epresentation \textbf{s}imilarity \textbf{m}atrix (RSM) \cite{kriegeskorte2008representational} obtained from model feature $\mathbf{z}$ for task $b$:
\begin{equation}
	\mathbf{M}_{ij}^{b} = \frac{(\mathbf{z}_i-\bar{\mathbf{z}})^T(\mathbf{z}_j-\bar{\mathbf{z}})}{||\mathbf{z}_i-\bar{\mathbf{z}}|| ||\mathbf{z}_j-\bar{\mathbf{z}}||},
	\label{rsmb}
\end{equation}
in which $\bar{\mathbf{z}}=\mathbb{E}_i[\mathbf{z}_i]$ denotes the mean model feature of all instances. Similarly, RSM obtained from affective dimension feature $\mathbf{\boldsymbol{\tau}}$ can be formulated as:
\begin{equation}
	\mathbf{M}_{ij}^{\text{aff}} = \frac{(\boldsymbol{\tau}_i-\bar{\boldsymbol{\tau}})^T(\boldsymbol{\tau}_j-\bar{\boldsymbol{\tau}})}{||\boldsymbol{\tau}_i-\bar{\boldsymbol{\tau}}|| ||\boldsymbol{\tau}_j-\bar{\boldsymbol{\tau}}||}.
	\label{rsmaff}
\end{equation}
Then, we define the similarity loss $\mathcal{L}_{kd_\text{aff}}$ as:
\begin{equation}
	\mathcal{L}_{kd_{\text{aff}}} = \mathbb{E}_{i\neq j}[\mathcal{L}_{kd_{\text{aff}}}^{ij}]
	= \mathbb{E}_{i\neq j}\left\lbrace [\mathrm{arctanh}(\mathbf{M}_{ij}^{b})-\mathrm{arctanh}(\mathbf{M}_{ij}^{\text{aff}})]^2\right\rbrace,
	\label{rkd_loss}
\end{equation}
in which $\mathcal{L}_{kd_{\text{aff}}}^{ij}$ is the sample based centered kernel alignment index. We leverage $\mathrm{arctanh}$ to reparameterize the similarity values from the interval $(-1,1)$ to $(-\infty,\infty)$ to approximately obey Gaussian distribution. Besides, to ensure training stability, we simultaneously pull together $\mathbf{M}_{ij}^{b}$ and $\mathbf{M}_{ij}^{b-1}$ using the same method to obtain $\mathcal{L}_{kd_{\text{model}}}$. In this way, the model has access to two teachers, affective dimension and the old model shown in Figure \ref{fig:kd}, which derives the overall knowledge distillation loss:
\begin{equation}
	\mathcal{L}_{kd} = \lambda_1 \mathcal{L}_{kd_{\text{model}}} + \lambda_2 \mathcal{L}_{kd_{\text{aff}}}.
	\label{final_loss}
\end{equation}

\subsection{Objective Function}
As mentioned above, the prediction confidence scores $\mathbf{s}$ for an instance $\mathbf{x}$ can be computed by Eq.\ref{confidence}, which denotes $\mathbf{s}=[s_1, \cdots, s_{|\mathcal{C}^b|}]^T\in \mathbb{R}^{|\mathcal{C}^b|}$ in task $b$. We have access to the ground truth from $\mathcal{D}^b$, which denotes as the multi-hot vector $\mathbf{y}=[y_1, \cdots, y_{C^b}]^T\in \mathbb{R}^{|C^b|}$. Besides, we have computed the soft label for previous emotion categories using the model $b-1$ shown in Section \ref{aERG}, which denotes as $\hat{\mathbf{s}}=[\hat{s}_1, \cdots, \hat{s}_{|\mathcal{C}^{b-1}|}]^T \in \mathbb{R}^{|\mathcal{C}^{b-1}|}$, and $C^b \cup \mathcal{C}^{b-1} = \mathcal{C}^{b}$. Above all, we train the task $b$ with the mixed ground truth $\tilde{\mathbf{y}} = [\hat{\mathbf{s}}^T,\mathbf{y}^T]^T \in \mathbb{R}^{|\mathcal{C}^b|}$using the binary cross entropy loss, formulated as:
\begin{equation}
	\mathcal{L}_{ce} = -\sum_{i=1}^{|\mathcal{C}^{b}|}[\tilde{y}_i\mathrm{log}(s_i)+(1-\tilde{y}_i)\mathrm{log}(1-s_i)].
\end{equation}
Finally, our model is trained with the following objective function in an end-to-end manner:
\begin{equation}
	\mathcal{L} = \mathcal{L}_{ce} + \lambda_1 \mathcal{L}_{kd_{\text{model}}} + \lambda_2 \mathcal{L}_{kd_{\text{aff}}} + \lambda_3 \mathcal{L}_{le}.
	\label{all}
\end{equation}
After training in task $b$, given an unseen instance, its associated label set is predicted as $\left\lbrace  \ell_k|s_k>0.5,1\leq k\leq \mathcal{C}^b \right\rbrace $.
\begin{figure*}[hbt!]
	\centering
	\subfigure[ \emph{Brain27}(1) ]{\includegraphics[scale=0.2]{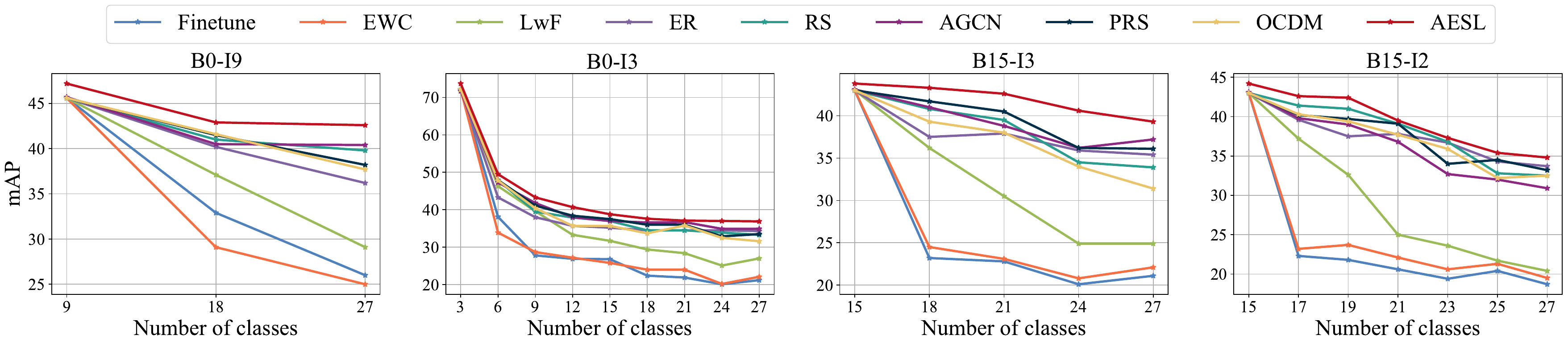}}
	\subfigure[ \emph{Video27} ]{\includegraphics[scale=0.2]{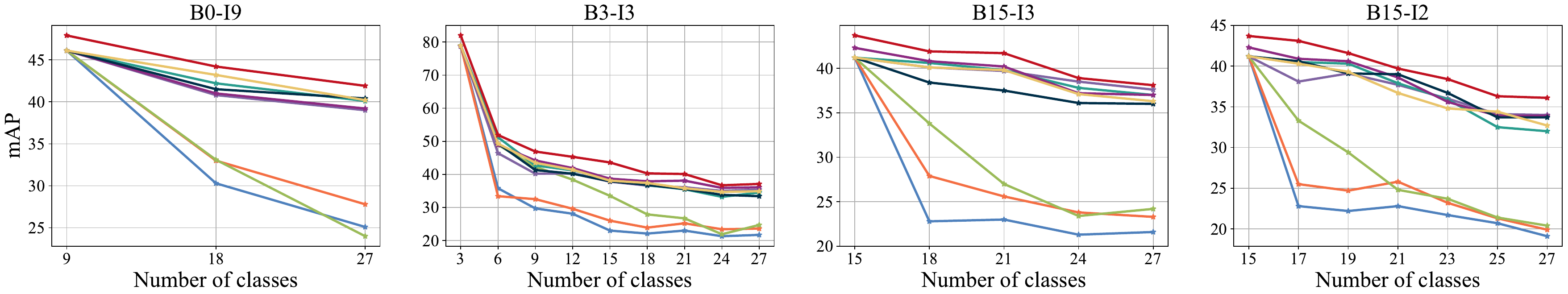}}
	\subfigure[ \emph{Audio28} ]{\includegraphics[scale=0.2]{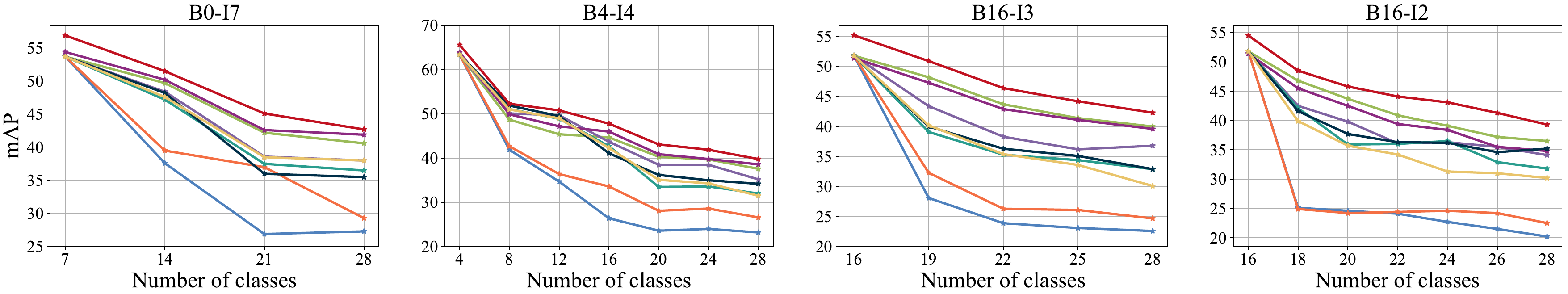}}
	\caption{Comparison results (mAP) on three datasets used in our experiment under different protocols against compared CIL methods.}
	\label{fig:multitask}
\end{figure*}

\section{Experiments}
\subsection{Experimental Setup}

\textbf{Datasets.}
For thoroughly evaluating the performance of AESL and comparing approaches, three datasets are leveraged for experimental studies including \emph{Brain27} \cite{horikawa2020neural}, \emph{Video27} \cite{cowen2017self} and \emph{Audio28} \cite{cowen2020music}. We evaluate our method using two popular protocols in class incremental learning work \cite{dong2023knowledge}, including (1) training all emotion classes in several splits and (2) first training a base model on a few classes while the remaining classes being divided into several tasks. For \emph{Brain27} and \emph{Video27}, we split the datasets with B0-I9 (base class is 0 and incremental class is 9), B0-I3, B15-I3 and B15-I2. For \emph{Audio28}, we split the dataset with B0-I7, B0-I4, B16-I3 and B16-I2. More details about the experiments can be found in Appendix \ref{datasets_d}.

\begin{table*}[h]
	\renewcommand\arraystretch{1.1}
	\scriptsize
	\caption{Class incremental results on subject 1 of \emph{Brain27} dataset. AGCN, PRS, OCDM are MLCIL algorithm in these compared methods.}
	\resizebox{\textwidth}{!}{
		\begin{tabular}{l|c|ccc|c|ccc|c|ccc|c|ccc}
			\hline
			\multirow{3}{*}{\textbf{Method}}  & \multicolumn{4}{c|}{\textbf{Brain27 B0-I9}}  & \multicolumn{4}{c|}{\textbf{Brain27 B0-I3}} & \multicolumn{4}{c|}{\textbf{Brain27 B15-I3}}&\multicolumn{4}{c}{\textbf{Brain27 B15-I2}} \\ \cline{2-17} 
			&    \multicolumn{1}{c|}{Avg. Acc} & \multicolumn{3}{c|}{Last Acc}& \multicolumn{1}{c|}{Avg. Acc} & \multicolumn{3}{c|}{Last Acc} &\multicolumn{1}{c|}{Avg. Acc} & \multicolumn{3}{c|}{Last Acc}&\multicolumn{1}{c|}{Avg. Acc} & \multicolumn{3}{c}{Last Acc}\\ \cline{2-17}  
			&   mAP &  maF1 & miF1 & mAP    & mAP  &  maF1 & miF1 & mAP & mAP  &  maF1 & miF1 & mAP& mAP  &  maF1 & miF1 & mAP  \\   
			\midrule 
			Upper-bound  &    - & 39.1  & 47.2  & 45.8 &    -   &   39.1  & 47.2  & 45.8 &  -&   39.1  & 47.2  & 45.8  & -  & 39.1  & 47.2  & 45.8 \\   
			\midrule
			Finetune  &   34.8  & 9.2  &19.3  &26.0  &  30.8 &5.0 &13.8  &21.2 &26.0  & 5.0  &13.9     &21.1 &23.7   & 3.6 &  13.2 & 18.7  \\ 
			EWC  &   33.2  & 8.1  &17.1  &25.0  & 30.9 &5.0 &13.9  &22.1 &  26.7&  5.3 & 14.3         &22.1  &  24.8 & 3.6 &  13.2 & 19.5 \\ 
			LwF  &  37.3  &12.2   &29.1  &29.1  & 37.0  &23.4 &40.0  &27.0 &  31.9& 14.2  &28.6       &24.9  &  29.1 &  15.2&  31.8 & 20.4 \\ 
			ER  &   40.7  & 8.0  &11.7  &36.2  &    40.2    &  4.9  &9.1  &34.3 & 37.9 &  9.7 &12.3 &  35.4   &  37.5&  9.6 & 11.7 &  33.7   \\ 
			RS  &  42.1   &  9.4 &12.6  &39.8  &    41.2    &  4.5  & 7.4 &33.3 &  38.3&  8.0 &  11.1& 33.9  &   37.9&  5.1 & 8.7 &   32.5 \\ 
			\midrule
			AGCN  &  42.2   & 29.5  &44.5  &40.4  &    42.1    &   35.4 & 43.7 &34.9 & 39.3 &28.8  & \textbf{41.7}& 37.2    & 36.3 & 24.4  & 36.4 &  30.9   \\ 
			PRS  &    41.6 &  9.3 &15.1  &38.2  &    41.7    &   5.5 & 8.4 & 33.5& 39.5 &  8.2 &   12.2&36.1  &  37.7& 9.4  & 13.1 &   33.2 \\ 
			OCDM  &     41.6& 9.7 & 15.9 & 37.7 &      40.6  &  5.3  &7.6  &31.6 & 37.2 &  4.9 &  7.1& 31.4  &  37.3&  4.4 & 7.6 &   32.5  \\ 
			\textbf{AESL}  & \textbf{44.2}   & \textbf{32.8}  &\textbf{44.7}  &\textbf{42.6} &\textbf{43.8} & \textbf{37.1} &\textbf{44.0}&\textbf{36.9}  & \textbf{41.9}&\textbf{32.5} &\textbf{41.7} & \textbf{39.3}  &\textbf{39.5}  & \textbf{26.8} & \textbf{36.5}  &\textbf{39.5}       \\ 
			\hline
		\end{tabular}
	}
	\label{tb:brain27}
\end{table*}

\begin{table*}[h]
	\renewcommand\arraystretch{1.1}
	\scriptsize
	\caption{Class incremental results on \emph{Video27} dataset. AGCN, PRS, OCDM are MLCIL algorithm in these compared methods.}
	\resizebox{\textwidth}{!}{
		\begin{tabular}{l|c|ccc|c|ccc|c|ccc|c|ccc}
			\hline
			\multirow{3}{*}{\textbf{Method}}  & \multicolumn{4}{c|}{\textbf{Video27 B0-I9}}  & \multicolumn{4}{c|}{\textbf{Video27 B0-I3}} & \multicolumn{4}{c|}{\textbf{Video27 B15-I3}}&\multicolumn{4}{c}{\textbf{Video27 B15-I2}} \\ \cline{2-17} 
			&    \multicolumn{1}{c|}{Avg. Acc} & \multicolumn{3}{c|}{Last Acc}& \multicolumn{1}{c|}{Avg. Acc} & \multicolumn{3}{c|}{Last Acc} &\multicolumn{1}{c|}{Avg. Acc} & \multicolumn{3}{c|}{Last Acc}&\multicolumn{1}{c|}{Avg. Acc} & \multicolumn{3}{c}{Last Acc}\\ \cline{2-17}  
			&   mAP &  maF1 & miF1 & mAP    & mAP  &  maF1 & miF1 & mAP & mAP  &  maF1 & miF1 & mAP& mAP  &  maF1 & miF1 & mAP  \\   
			\midrule 
			Upper-bound  &    - &  36.8 & 46.3  & 45.4 &    -   &    36.8 & 46.3  & 45.4  &  -&    36.8 & 46.3  & 45.4   & -  &  36.8 & 46.3  & 45.4  \\   
			\midrule
			Finetune  & 33.8  & 6.6 & 13.7 & 25.1 & 31.5 &4.2 & 13.0 &21.7 & 26.0 &  4.3 & 13.3    &24.3 & 21.6&4.2  & 13.7 & 19.1     \\ 
			EWC  &   35.6  &  7.5 &  17.8&  27.8& 32.9 &   4.7 & 13.2 & 23.6& 28.4 &  4.9 &  13.9   & 26.0 & 23.3& 3.9 & 13.1 & 19.9    \\ 
			LwF  &   34.4  &  6.8 &  23.3& 24.0 & 38.2  &  19.9  & 37.3 & 24.7& 30.0 &  12.5 & 33.4 & 27.7 &24.2&  15.8 & 32.7 & 20.4    \\ 
			ER  &    42.0 &  5.1 &  7.0& 39.0 &   43.0 &  4.5  & 4.8 & 35.4& 39.4 & 8.0  &   8.3  & 37.2 & 37.6& 10.1 & 12.6 & 34.0    \\ 
			RS  &   42.8  &  4.6 & 6.1 & 40.1 &  43.6 &   4.6 & 7.8 & 34.5& 39.3 &  4.2 &  5.2   &  37.2 & 37.0& 6.7 & 9.9 & 32.0  \\ 
			\midrule
			AGCN  &  42.1   &  22.4 & 39.4 & 39.2 &  44.5 &34.2    & 44.5 &36.1 & 39.5 & 22.7  & 38.4   & 38.0 &37.0& 23.8  &36.2  & 34.0    \\ 
			PRS  &   42.6  &  9.5 &  15.0& 40.4 &    43.0&   5.8 & 9.6 & 33.4&  37.8&  8.9 &   13.4  & 37.7 & 36.0& 7.4 & 13.3 & 33.7    \\ 
			OCDM  &    43.1 &  5.5 & 6.8 & 40.2 &     43.8&   5.0 &  7.8&35.0 & 38.9 &  4.9 &   6.6  &  37.1& 36.3& 5.2 & 5.8 &  32.7   \\ 
			\textbf{AESL}  & \textbf{44.6}   &\textbf{23.4}   &\textbf{39.7}  & \textbf{41.9} & \textbf{47.1} & \textbf{35.2}   & \textbf{45.0} & \textbf{37.1}&\textbf{41.5}  & \textbf{23.5}  & \textbf{39.2} &\textbf{39.8}  &\textbf{38.1}&  \textbf{24.5}  & \textbf{36.7} & \textbf{36.1}   \\ 
			\hline
		\end{tabular}
}
	\label{tb:video27}
\end{table*}

\begin{table*}[h]
	\renewcommand\arraystretch{1.1}
	\scriptsize
	\caption{Class incremental results on \emph{Audio28} dataset. AGCN, PRS, OCDM are MLCIL algorithm in these compared methods.}
	\resizebox{\textwidth}{!}{
		\begin{tabular}{l|c|ccc|c|ccc|c|ccc|c|ccc}
			\hline
			\multirow{3}{*}{\textbf{Method}}  & \multicolumn{4}{c|}{\textbf{Audio28 B0-I7}}  & \multicolumn{4}{c|}{\textbf{Audio28 B0-I4}} & \multicolumn{4}{c|}{\textbf{Audio28 B16-I3}}&\multicolumn{4}{c}{\textbf{Audio28 B16-I2}} \\ \cline{2-17} 
			&    \multicolumn{1}{c|}{Avg. Acc} & \multicolumn{3}{c|}{Last Acc}& \multicolumn{1}{c|}{Avg. Acc} & \multicolumn{3}{c|}{Last Acc} &\multicolumn{1}{c|}{Avg. Acc} & \multicolumn{3}{c|}{Last Acc}&\multicolumn{1}{c|}{Avg. Acc} & \multicolumn{3}{c}{Last Acc}\\ \cline{2-17}  
			&   mAP &  maF1 & miF1 & mAP    & mAP  &  maF1 & miF1 & mAP & mAP  &  maF1 & miF1 & mAP& mAP  &  maF1 & miF1 & mAP  \\   
			\midrule 
			Upper-bound  &    - & 51.4  &61.1   &57.1  &    -   &   51.4  &61.1   &57.1 &  -&    51.4  &61.1   &57.1  & -  &  51.4  &61.1   &57.1 \\   
			\midrule
			Finetune  &  36.4   & 9.2 & 14.8&27.3 & 33.9 & 5.3 & 10.0& 23.3 & 29.9& 4.4 & 10.3  &   22.6  &27.6 & 2.8  & 8.2 & 20.2     \\ 
			EWC  &   37.9  &  8.3 & 14.3 &29.3&37.1  &  5.4  &   10.5 & 26.6 & 32.2& 4.4 &  9.7 &   24.7  & 28.1 &   2.8& 8.7 & 22.5    \\ 
			LwF  &    46.6 &  37.9 & 51.7 &40.6& 45.8 &   39.9 &   49.8 & 37.6 & 45.0& \textbf{32.3} & 45.2  &  40.0   &42.3  &28.8   &41.4  & 36.5    \\ 
			ER  &    44.7 &   8.1& 11.4 &38.0& 45.6 &    6.5&   5.5 & 35.2 &41.3 & 9.2 & 13.3  &   36.8  & 39.4 &  10.1 & 13.6 & 34.1    \\ 
			RS  &    43.7 &   8.1& 12.3&36.5 & 43.6 &    5.9 &  9.3  &  32.0& 38.7& 7.5 &  11.7 &   32.9  & 38.2  &  5.8 & 11.6 & 31.8   \\ 
			\midrule
			AGCN  &  47.3   & 35.3  & 50.9&41.9 &46.6  &   37.5    &51.0  &38.6  &44.5 &29.3  & 44.6  & 39.6 & 41.1 & 27.5  & 42.4 & 34.8    \\ 
			PRS  &    43.3 &  9.0 & 12.8 &35.5& 44.5 &     6.8&   9.0 & 34.2 & 39.2& 6.2 &  8.3 &   32.9  & 39.1 &  8.8 & 11.4 & 35.2    \\ 
			OCDM  &    44.5 &  8.7 & 12.0 &38.0 & 43.8 &    7.5&   8.8 & 31.5 & 38.2& 5.5 &  9.7 &    30.1 & 36.3 &  3.7 & 7.9 &30.2     \\ 
			\textbf{AESL}  &\textbf{49.0}     &\textbf{38.4}   & \textbf{51.8}&\textbf{42.7} &\textbf{48.7}  & \textbf{41.1}  & \textbf{51.7}   & \textbf{39.8} &\textbf{47.8} &\textbf{32.3}  &\textbf{48.0}   &  \textbf{42.3}  & \textbf{45.3} & \textbf{30.8}  &\textbf{45.1}  &\textbf{39.3}     \\ 
			\hline
		\end{tabular}
	}
	\label{tb:audio28}
\end{table*}

\subsection{Experimental Results}
\textbf{Comparative Studies.}
\label{comparative studies}
Tables \ref{tb:brain27}, \ref{tb:video27} and \ref{tb:audio28} show the results on subject 1 of \emph{Brain27} (more results are shown in Appendix \ref{more_result}), \emph{Video27} and \emph{Audio}, respectively. We can observe that AESL shows obvious superiority under different datasets and protocols, in terms of three widely used metrics mAP, maF1 and miF1 \cite{zhang2013review}. Especially for \emph{Brain27}, AESL has a relative improvement of 9.6\% on mAP (4 protocols averaged) and 9.7\% on maF1 compared with the second place method. Figure \ref{fig:multitask} exhibits the comparison curves of AESL and comparing methods, which indicates that our method is consistently optimal at each task of incremental learning.
Among the compared methods, EWC and LwF are traditional SLCIL methods. We can find that EWC is not suitable for direct application to MLCIL task due to its lowest performance. In contrast, LwF achieves impressive performance on the \emph{Audio28} dataset. We infer that knowledge distillation, which essentially provides soft labels for old classes, contributes to overcoming catastrophic forgetting in MLCIL. Besides, it is noticeable that rehearsal-based methods, including ER, RS, PRS and OCDM, are not satisfactory, especially for maF1 and miF1. This is because just saving the labels of current task aggravates the partial label problem in subsequent training, which illustrates that data replay is not suitable for MLCIL. Furthermore, AGCN is a strong baseline which can rank 2nd for most cases, and our method suppresses AGCN by incorporating the effective label semantics learning and introducing knowledge distillation from affective space.

\begin{table}[h]
	\renewcommand\arraystretch{1}
	\caption{The contribution of each component. Accuracy of these models is measured by mAP.}
	\begin{center}
		\begin{tabular}{l|ccc|cc}
			\hline
			Model & ESL & LD & RKD & Avg. Acc & Last Acc   \\
			\midrule
			w/o ESL &   &   & \checkmark & 47.6  & 41.3  \\
			w/o LD & \checkmark  &   & \checkmark  &48.1    &42.0     \\
			w/o RKD & \checkmark & \checkmark  & & 48.3  &42.1     \\ 
			\midrule
			\textbf{AESL}  & \checkmark & \checkmark & \checkmark & \textbf{49.0}  &\textbf{42.7} \\
			\hline
		\end{tabular}
	\end{center}
	\label{tb:abalation}
\end{table}

\textbf{Ablation Studies.}
In order to test the role of three key components of AESL separately, we conduct the ablation experiment on the \emph{Video28} dataset under B0-I7 setting. We design three baselines whose results are shown in Table \ref{tb:abalation}. (1) \textbf{w/o ESL}: Feed the feature vector \textbf{z} directly into the classifier without semantic-guided feature decoupling. (2) \textbf{w/o LD}: Use original confidence score $\textbf{S}$ for constructing adjacency matrix without label disambiguation. (3) \textbf{w/o RKD}: Remove the module of knowledge distillation from affective space. Results show that all the three components in AESL are critical for preventing forgetting and improving the model performance of MLCIL.

\begin{figure}
	\centering
	\subfigure[ \emph{Brain27} ]{\includegraphics[scale=0.45]{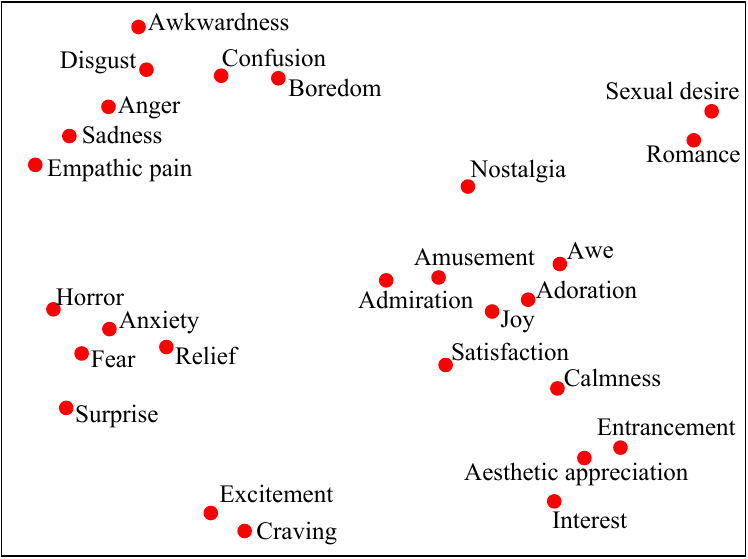}}
	\hspace{0.3cm}
	\subfigure[ \emph{Audio28} ]{\includegraphics[scale=0.45]{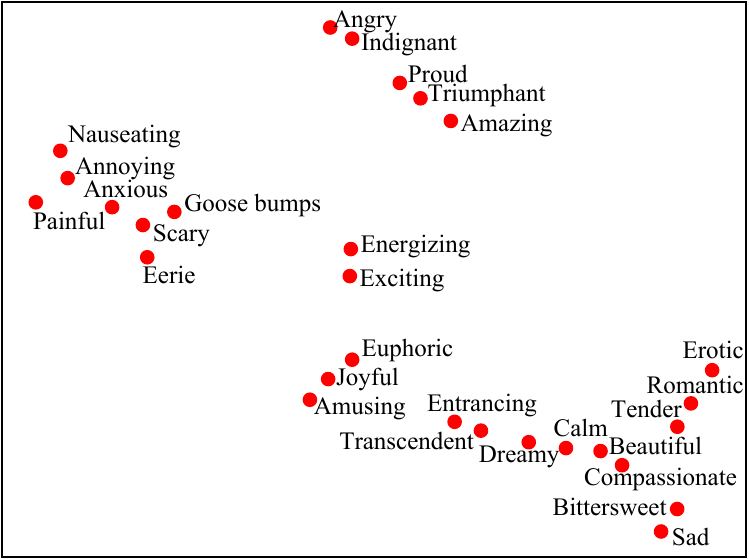}}
	\caption{t-SNE on the emotion embeddings learning by AESL.}
	\label{fig:tsne}
\end{figure}

\textbf{Emotional Semantics Visualization.}
Emotional semantics learning is an essential module for our approach. In Figure \ref{fig:tsne}, we adopt the t-SNE \cite{van2008visualizing} to visualize the emotion embeddings learned by the emotional semantics learning module. It is clear to see that, the learned embeddings maintain meaningful emotional semantic topology. Specifically, in \emph{Brain27}, positive emotions are predominantly distributed in the bottom right, while negative emotions are distributed in the top left. In \emph{Audio28}, \emph{Bittersweet} is just located between \emph{Sad} and other positive emotions. This visualization further demonstrates the necessity of modeling label dependencies.

\begin{figure}[h]
	\begin{center}
		\includegraphics[scale=0.45]{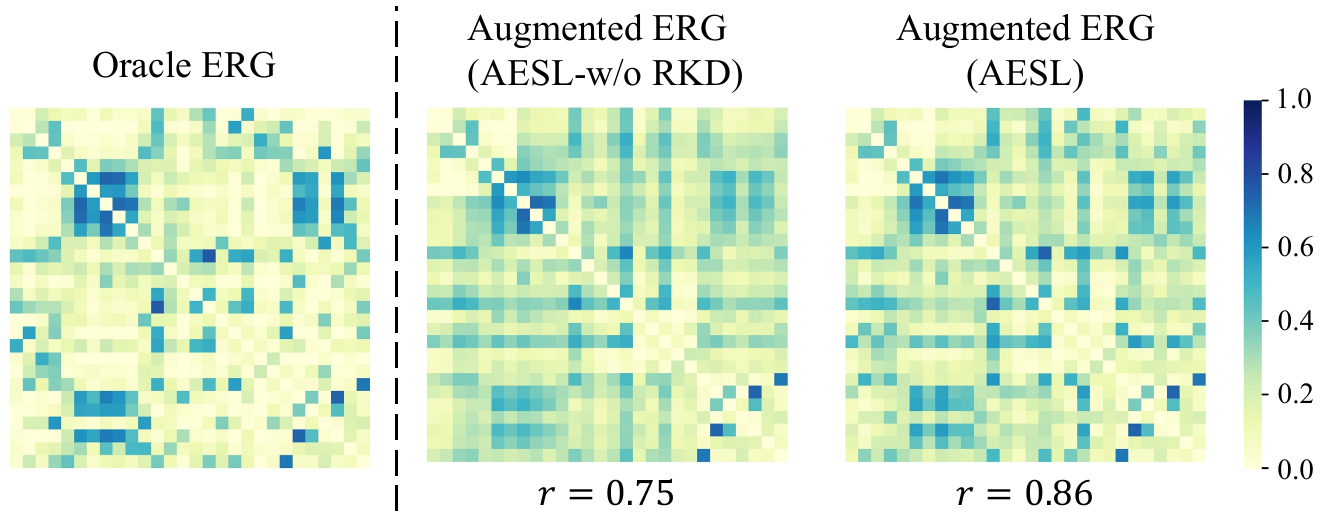}
	\end{center}
	\caption{Visualization of augmented emotional relation graph in \emph{Audio28} dataset.}
	\label{fig:label_adj}
\end{figure}

\textbf{Augmented ERG Visualization.}
In Figure \ref{fig:label_adj}, we provide the augmented ERG visulization on \emph{Audio28} dataset. We utilize the oracle ERG, which is constructed using ground truth label statistics of all tasks, as the upper bound. We also compute the \textbf{P}earson's \textbf{C}orrelation \textbf{C}oefficients (PCCs) to measure the similarity between the augmented and oracle ERG. We can observe that using the proposed method, inter- and intra-task label relationships are reconstructed well. Besides, by incorporating RKD from affective dimension space, augmented ERG is closer to oracle ERG ($r=0.86$ vs $r=0.75$). We speculate that this is due to the introduction of RKD, which is conducive to alleviating the issue of future-missing partial label.

\textbf{Parameter Sensitivity.}

\begin{figure}
	\begin{center}
		\includegraphics[scale=0.4]{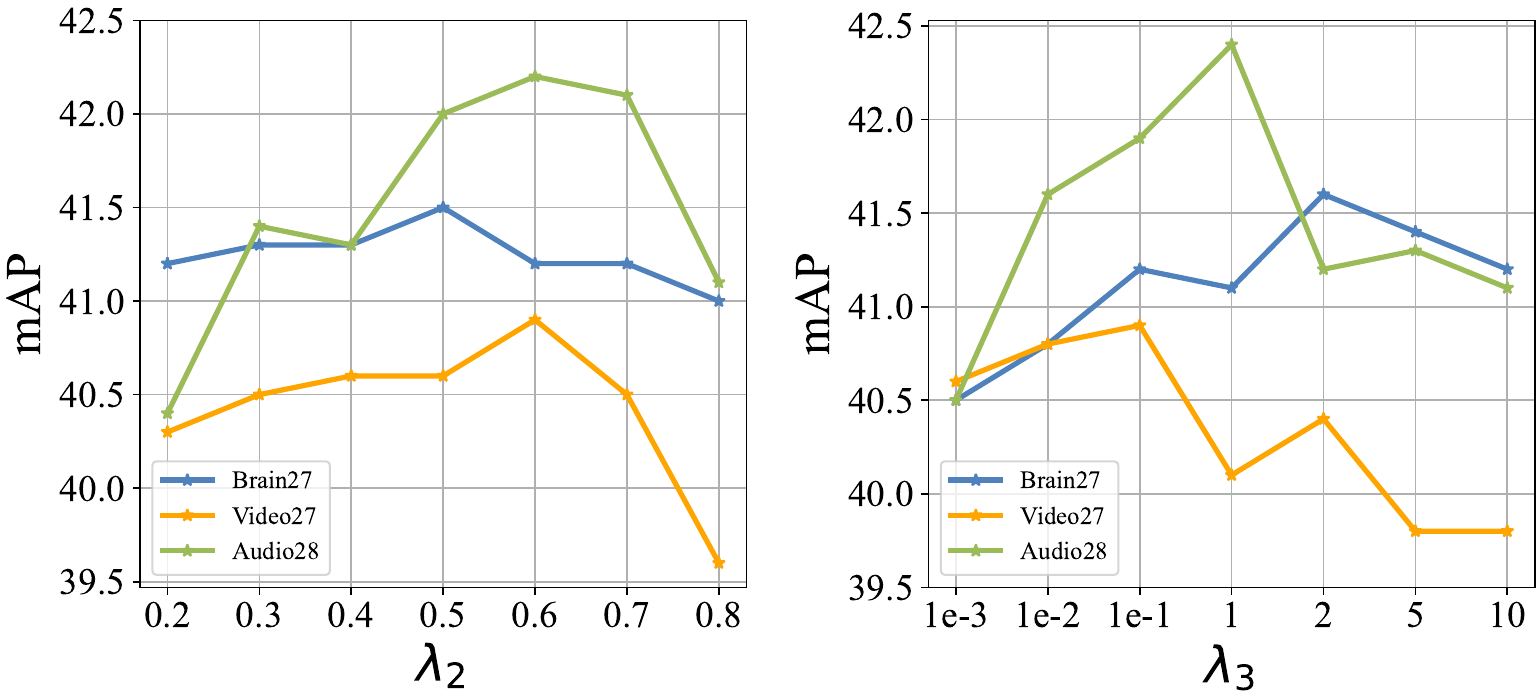}
	\end{center}
	\caption{The performance of AESL measured by last mAP changes as $\lambda_2$ and $\lambda_3$ vary.}
	\label{fig:sen}
\end{figure}

Figure \ref{fig:sen} gives an illustrative example of how the performance of AESL changes as the regulation parameter $\lambda_2$ and $\lambda_3$ vary on the B0-I9 and B0-I7 protocols of three datasets ($\lambda_1$ is fixed to 1). Here, when the value of one parameter varies, the other is fixed to a reasonable value. We find that too large value of $\lambda_2$ dramatically degrade the model performance due to the noise in affective ratings, while too small value will not play the role of alleviating the future-missing partial label problem. As for $\lambda_3$, too large value will cause the model to focus too much on graph reconstruction, while too small value will not be able to learn label embeddings well, which both lead to the descend of model performance.

\section{Conclusion}
In this paper, we have proposed a novel AESL framework for multi-label class incremental emotion decoding. 
In detail, we developed an augmented ERG generation method with label disambiguation for handling the past-missing partial label problem.
Then, knowledge distillation from affective dimension space was introduced for alleviating future-missing partial label problem.
Besides, we constructed an emotional semantics learning module to learn indispensable label embeddings for subsequent semantic-specific feature extraction.
Extensive experiments have illustrated the effectiveness of AESL.

\bibliographystyle{unsrtnat}
\bibliography{neurips_2024.bib}

\newpage
\appendix
\section{Related Work}
\textbf{Class Incremental Emotion Decoding.}
In the last few years, continuous learning of new emotion categories has been increasingly studied. \cite{churamani2020clifer} proposed CLIFER, which combined a generative model and a complementary learning-based dual-memory model, to enable continual facial expression recognition. Besides, \cite{ma2022few} presented a graph convolutional networks (GCN) based method for few-shot class-incremental classification within five emotion categories. \cite{jimenez2022class} performed a weight aligning to correct the bias on new emotion classes. However, these studies are limited to a small number of coarse-grained emotion categories and fail to consider the complexity of human emotion expression.

\textbf{Multi-label Learning.}
The core issue in multi-label learning is modelling the dependencies between labels. There exists some studies which utilized GCN for label specific feature learning \cite{chen2019learning,chen2019multi}. Some approaches have also leveraged Transformer to capture the relationships between instances and labels \cite{zhao2021transformer,liu2021query2label}.
Regarding multi-label emotion decoding, \cite{fei2020latent} proposed latent emotion memory (LEM) for learning latent emotion distribution and utilized bi-directional GRU to learn emotion coherence based on text data. \cite{fu2022multi} put forward a multi-view multi-label hybrid model for emotion decoding from human brain activity. Nevertheless, these models lack the capability of continual learning.

\textbf{Multi-label Class Incremental Learning.}
In many real-world application scenarios, it is essential for models to be capable of incrementally learning new classes and achieving multi-label classification simultaneously. \cite{dong2023knowledge} proposed a knowledge restore and transfer framework based on attention module. AGCN \cite{du2022agcn} constructed the relationship between labels and leveraged GCN to learn them. PRS \cite{kim2020imbalanced} and OCDM \cite{liang2022optimizing} achieved multi-label online class incremental learning by implementing specialized design for replay buffer. However, MLCIL focused on emotion decoding is less-studied in the literature.

\section{Details of Experiments}
\label{datasets_d}

\subsection{Details of Datasets}
Table \ref{data sets} shows the characteristics of the three datasets used in our experiments. Properties of each dataset are characterized by several statistics, including number of training instances $|\mathcal{D}_{tr}|$, number of test instances $|\mathcal{D}_{te}|$, number of features $Dim(\mathcal{D})$, threhold for constructing label marix $Th(\mathcal{D})$, number of possible class labels $L(\mathcal{D})$, number of affective dimensions $Aff(\mathcal{D})$, label cardinality (average number of labels per instance) $LCard(\mathcal{D})$, label density (label cardinality over $L(\mathcal{D})$) $LDen(\mathcal{D})$ and modality. For $\emph{Brain27}$, we also exhibit the number of voxels $V(\mathcal{D})$ before ROI-pooling used in our experiments. Below, we will provide some details about these datasets.
\begin{table*}
	\renewcommand\arraystretch{1.2}
	\small
	\centering
	\caption{The characteristics of the experimental datasets.}
	\label{data sets}
	\resizebox{\textwidth}{!}{
	\begin{tabular}{c c c c c c c c ccc}
		\hline
		Dataset           & $|\mathcal{D}_{tr}|$    & $|\mathcal{D}_{te}|$   &$V(\mathcal{D})$      &  $Dim(\mathcal{D})$    & $Th(\mathcal{D})$ & $L(\mathcal{D})$ &$Aff(\mathcal{D})$ & $LCard(\mathcal{D})$    & $LDen(\mathcal{D})$&  Modality            \\ 
		\hline
		\emph{Brain27}(1)        & 1800               & 396     &  120930   & 2880   &0.10     &27 &14 &4.64  & 0.17 &fMRI   \\
		\emph{Brain27}(2)        & 1800               & 396     &  116260   & 2880   &0.10     &27 &14 &4.64  & 0.17 &fMRI   \\
		\emph{Brain27}(3)        & 1800               & 396     &  102941   & 2880   &0.10     &27 &14 &4.64  & 0.17 &fMRI   \\
		\emph{Brain27}(4)        & 1800               & 396     &  118533   & 2880   &0.10     &27 &14 &4.64  & 0.17 &fMRI   \\
		\emph{Brain27}(5)        & 1800               & 396     &  116699   & 2880   &0.10     &27 &14 &4.64  & 0.17 &fMRI   \\
		\emph{Video27}        & 1800               & 396      &  -  & 1000   &0.10    &27 & 11 &4.64 & 0.17 &Video  \\
		\emph{Audio28}        & 1500               & 341   &  -     & 512    &0.15    &28  &11 &  5.27 &  0.19  &Audio  \\
		\hline
	\end{tabular}}
\end{table*}

\emph{Brain27} is a visually evoked emotional brain activity dataset \cite{horikawa2020neural}, which contains the blood-oxygen-level dependent (BOLD) responses of five subjects, who were shown 2196 video clips while functional Magnetic Resonance Imaging (fMRI) data were recorded. These data were collected using a 3T Siemens scanner with a multiband gradient Echo-Planar Imaging (EPI) sequence (TR, 2000ms; TE, 43ms; flip angle, 80 deg; FOV, 192$\times$192 mm; voxel size, 2$\times$2$\times$2 mm; number of slices, 76; multiband factor, 4). The fMRI data was preprocessed and averaged with each video stimulus, which means the brain activity of one voxel is a scalar for a video stimulus.

\emph{Video27} is an emotionally evocative visual dataset \cite{cowen2017self}, which had been used to collect the brain activity in \emph{Brain27}. This dataset contains 2196 videos whose durations ranged from 0.15s to 90s. Some video screenshots of \emph{Video27} has been shown in Figure \ref{fig:mi}.

\emph{Audio28} is an emotionally evocative auditory dataset \cite{cowen2020music}, which consists of 1841 music samples without lyrics. In these music clips, 1572 were selected from YouTube, 88 came from Howard Shore's \emph{Lord of the Rings} soundtrack, and 181 came from Wagner's \emph{Ring} cycle. These segments of music can convey strong feelings, whose durations ranged from 0.73s to 7.89s.

In terms of emotion category an affective dimension ratings, in \emph{Brain27} and \emph{Video27}, each instance was voted by multiple raters across 27 emotion categories and 14 affective dimensions. In \emph{Audio28}, each music clip was judged by multiple rates across 28 emotion categories and 11 affective dimensions. Emotion category ratings range from 0 to 1 and we set threshold 0.1 for \emph{Brain27} and \emph{Video27} and 0.15 for \emph{Audio28} to construct emotion label matrix. The average number of emotion labels for the former two datasets and \emph{Audio28} is 4.64 and 5.27, respectively. Affective dimension ratings were rated by 9-scale Likert scale, which are standardized before RKD in our experiments.

In the process of splitting emotion labels for incremental learning,  we just follow the order of the alphabet without other interfere. The order of \emph{Brain27} and \emph{Video27} is \emph{Admiration}, \emph{Adoration}, \emph{Aesthetic appreciation}, \emph{Amusement}, \emph{Anger}, \emph{Anxiety}, \emph{Awe}, \emph{Awkwardness}, \emph{Boredom}, \emph{Calmness}, \emph{Confusion}, \emph{Craving}, \emph{Disgust},\emph{Empathic pain}, \emph{Entrancement},\emph{Excitement}, \emph{Fear}, \emph{Horror}, \emph{Interest}, \emph{Joy}, \emph{Nostalgia}, \emph{Relief}, \emph{Romance}, \emph{Sadness}, \emph{Satisfaction}, \emph{Sexual desire} and \emph{Surprise}. The order of \emph{Audio28} is \emph{Amusing}, \emph{Angry}, \emph{Annoying}, \emph{Anxious}, \emph{Amazing}, \emph{Beautiful}, \emph{Bittersweet}, \emph{Calm}, \emph{Compassionate}, \emph{Dreamy}, \emph{Eerie}, \emph{Energizing}, \emph{Entrancing}, \emph{Erotic}, \emph{Euphoric}, \emph{Exciting}, \emph{Goose bumps}, \emph{Indignant}, \emph{Joyful}, \emph{Nauseating}, \emph{Painful}, \emph{Proud}, \emph{Romantic}, \emph{Sad}, \emph{Scary}, \emph{Tender}, \emph{Transcendent} and \emph{Triumphant}. 

Affective dimensions used in \emph{Brain27} and \emph{Video27} are \emph{Approach}, \emph{Arousal}, \emph{Attention}, \emph{Certainty}, \emph{Commitment}, \emph{Control}, \emph{Dominance}, \emph{Effort}, \emph{Fairness}, \emph{Identity}, \emph{Obstruction}, \emph{Safety}, \emph{Upswing} and \emph{Valence}. Affective dimensions used in \emph{Audio28} are \emph{Arousal}, \emph{Attention}, \emph{Certainty}, \emph{Commitment}, \emph{Dominance}, \emph{Enjoyment}, \emph{Familiarity}, \emph{Identity}, \emph{Obstruction}, \emph{Safety} and \emph{Valence}.

\subsection{Comparing Approaches}
The performance of  AESL is compared with multiple essential and state-of-art class incremental methods. \emph{Finetune} is a baseline which means fine-tuning the model without any anti-forgetting constraints. We select four SLCIL methods including \emph{EWC} \cite{kirkpatrick2017overcoming}, \emph{LwF} \cite{lee2019overcoming}, \emph{ER} \cite{rolnick2019experience} and \emph{RS} \cite{vitter1985random} for comparison. Furthermore, other three well-established MLCIL approaches \emph{AGCN} \cite{du2022agcn}, \emph{PRS} \cite{kim2020imbalanced} and \emph{OCDM} \cite{liang2022optimizing} are also employed as comparing approaches. Besides, we set the \emph{Upper-bound} as the supervised training on the data of all tasks. Details of these compared methods are as follows.

\textbf{EWC}\cite{kirkpatrick2017overcoming}: A Single-Label Class Incremental Learning algorithm that reduces catastrophic forgetting by constraining important parameters uses the Fisher information matrix to compute the importance of parameters.

\textbf{LWF}\cite{lee2019overcoming}: The first algorithm to apply knowledge distillation to the Single-Label Class Incremental Learning task uses the old model as a teacher and minimizes the KL divergence between the probability distributions of the outputs of the new and old models.

\textbf{ER}\cite{lee2019overcoming}: A Single-Label Class Incremental Learning algorithm based on data replay, where the construction of a data buffer employs a random sampling strategy.

\textbf{RS}\cite{vitter1985random}: A Single-Label Class Incremental Learning algorithm based on data replay, where the construction of a data buffer utilizes a reservoir sampling strategy.

\textbf{AGCN}\cite{vitter1985random}: A Multi-Label Class Incremental Learning algorithm based on graph convolutional neural networks, where the graph adjacency matrix continuously expands as the tasks progress.

\textbf{PRS}\cite{kim2020imbalanced}: A Multi-Label Class Incremental Learning algorithm based on data replay, which improves upon the reservoir sampling strategy to ensure that the number of samples for each class in the data buffer is as balanced as possible.

\textbf{OCDM}\cite{liang2022optimizing}: A Multi-Label Class Incremental Learning algorithm based on data replay that defines the construction and updating of the data buffer as an optimization problem to be solved.

\subsection{Feature Extraction}
In \emph{Brain27} dataset, we extract 2880-dimensional feature vector with ROI-pooling (see Appendix \ref{roi_p}). In \emph{Video27}, visual object features have been extracted with pre-trained VGG19 model \cite{simonyan2014very} for one frame and averaged across all frames to construct 1000-dimensional features. In \emph{Audio28} dataset, we compute \textbf{M}el-\textbf{f}requency \textbf{c}epstral \textbf{c}oefficients (MFCC) for each audio fragment. All MFCC fragments from the same audio are then input into pre-trained ResNet-18 \cite{he2016deep} model and averaged across all fragments to obtain 512-dimensional features.

\subsection{Hyperparameters Settings}
In our experiments, the balancing parameter $\beta$ is set to 0.95 in Eq.\ref{pro}. We set $\lambda_1$ to 1 in Eq.\ref{all}. Besides, $\lambda_2$ is searched in $\left\lbrace0.2, 0.3, 0.4, 0.5, 0.6, 0.7, 0.8 \right\rbrace$ and $\lambda_3$ is searched in $\left\lbrace 0.001, 0.01, 0.1, 1, 2, 5, 10 \right\rbrace$. The dimensionality of deep latent representations $\mathbf{z}$ is set to 64 in three datasets. We train the model using the Adam optimizer with $\left\lbrace \beta_1,\beta_2 \right\rbrace = \left\lbrace 0.9,0.9999 \right\rbrace$. We set weight decay of 0.005 and learning rate of $10^{-4}$ for \emph{Brain27} and \emph{Video27}, and weight decay of 0 and learning rate of $10^{-3}$ for \emph{Audio28}. We conducted all the experiments on one NVIDIA TITAN GPU.

\section{Details of ROI-Pooling}
\label{roi_p}
\begin{figure}
	\centering
	\includegraphics[scale=0.65]{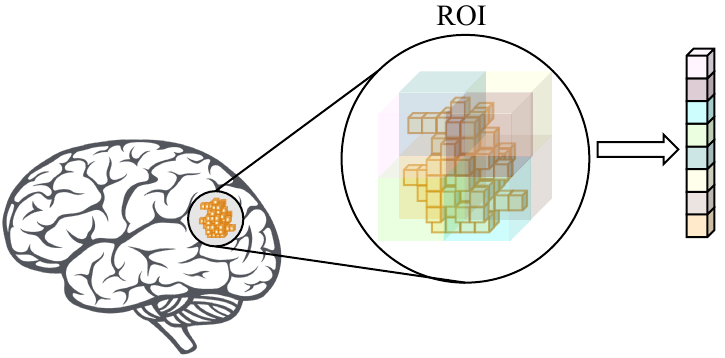}
	\caption{A schematic diagram of ROI pooling. Each orange small cube represents a voxel.}
	\label{fig:roi}
\end{figure}

Using the brain voxels signal directly for voxel-wise decoding will introduce lots of noise and easily cause overfitting. Therefore, we first use the HCP360 template \cite{glasser2016multi} to divide the whole brain into multiple brain regions (ROIs) that include 360 cortical regions defined by a parcellation provided from the Human Connectome Project. In order to further extract the features of each ROI, we place the voxels of each ROI in a 3-D volume according to its coordinates, then split the volume evenly into 8 sub-volumes and calculate the average brain activity of voxels in each sub-volume as the feature of this sub-volume as illustrated in Figure \ref{fig:roi}. In other words, for each ROI we get 8-dimensional features. Then we concat the features of 360 ROIs in each hemisphere to get 2880-dimensional features.
\section{The Algorithm of AESL}
\label{algorithm}
The algorithm of AESL is presented in Algorithm \ref{alg:AESL}.
\section{More Comparative Results}
\label{more_result}
\begin{figure*}[hbt!]
	\centering
	\subfigure[ \emph{Brain27}(2) ]{\includegraphics[scale=0.23]{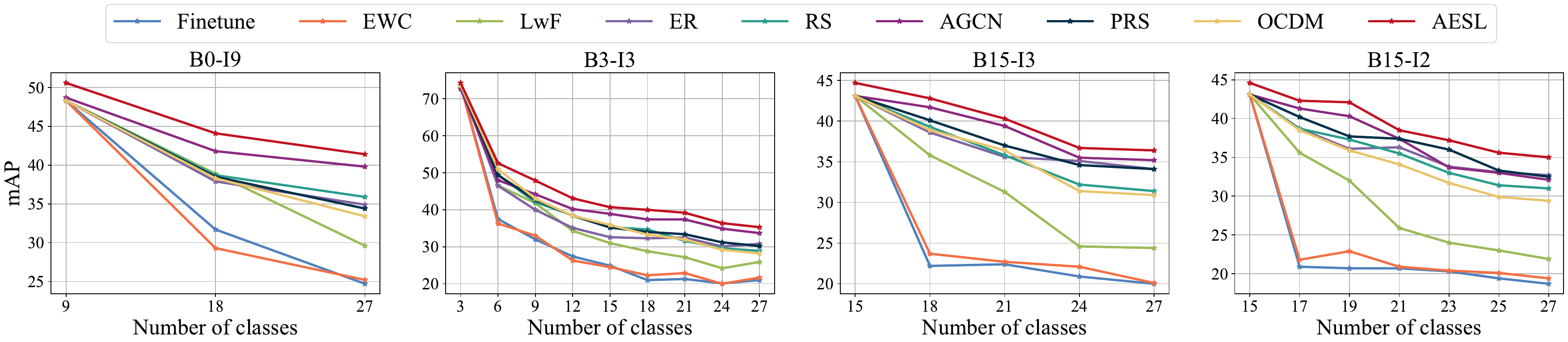}}
	\subfigure[ \emph{Brain27}(3) ]{\includegraphics[scale=0.23]{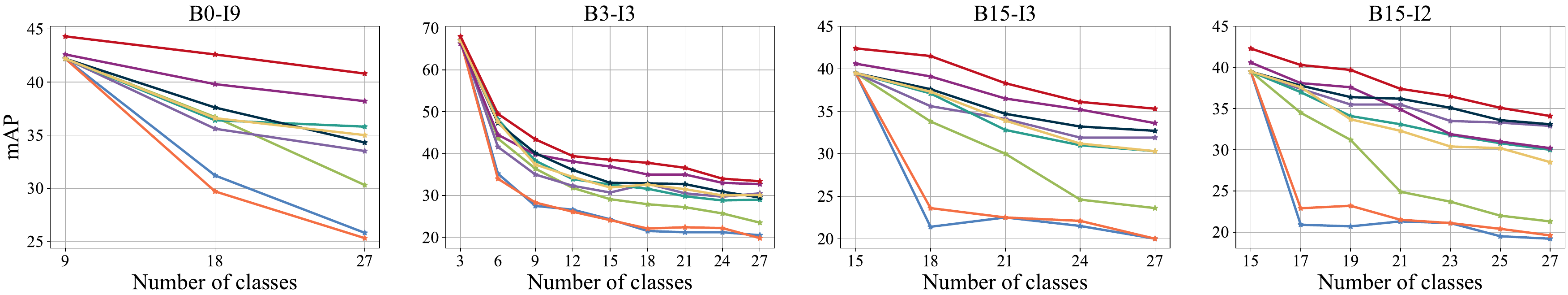}}
	\subfigure[ \emph{Brain27}(4) ]{\includegraphics[scale=0.23]{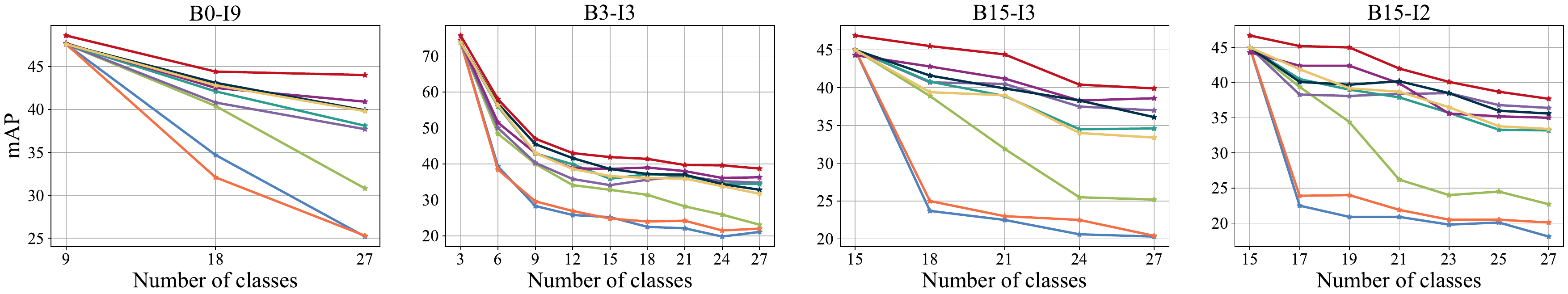}}
	\subfigure[ \emph{Brain27}(5) ]{\includegraphics[scale=0.23]{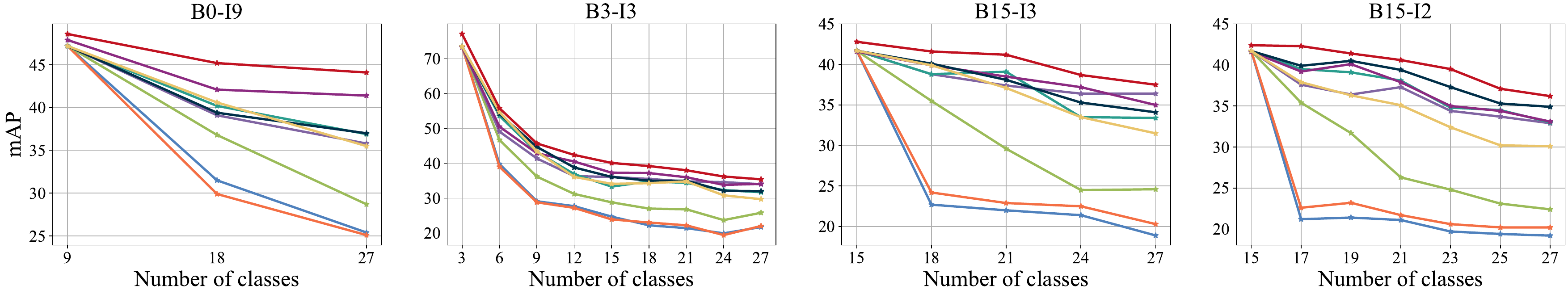}}
	\caption{Comparison results (mAP) on three datasets used in our experiment under different protocols against compared CIL methods.}
	\label{fig:multitask_app}
\end{figure*}

Tables \ref{tb:brain27(2)}, \ref{tb:brain27(3)}, \ref{tb:brain27(4)} and \ref{tb:brain27(5)} show the results on subject 2, subject 3, subject4 and subject 5 in \emph{Brain27} dataset. Figure \ref{fig:multitask_app} exhibits the comparison curves of AESL and comparing methods for these subjects in \emph{Brain27} dataset. We observe that similar conclusions can be drawn as mentioned in Section \ref{comparative studies}.

\begin{table}
	\centering
	\renewcommand\arraystretch{1.2}
	\setlength\tabcolsep{12pt}
	\caption{Friedman statistics $F_F$ in terms of each metric and the critical value at 0.05 significance level. ($\#$ compared algorithms $k=9$, $\#$ subjects $N=7$.)}
	\begin{tabular}{ccc|c} 
		\hline
		maF1 & miF1& mAP& critical value\\
		\hline	
		37.025 &64.000& 70.696& 2.138\\
		\hline
	\end{tabular}
	\label{friedman}
\end{table}

Furthermore, we adopt the Friedman test \cite{demvsar2006statistical} for statistical test in order to discuss the relative performance among the compared methods\footnote{We average the last mAP of four protocols regarding each dataset for further analysis.}. If there are $k$ algorithms and $N$ datasets, we take use of the average ranks of algorithms $R_j = \frac{1}{N}\sum_{i}r_i^j$ for Friedman test in which $r_i^j$ is the ranks of the $j$-th algorithm on the $i$-th dataset. If the null-hypothesis is that all the algorithms have the equivalent performance, the Friedman statistic $F_F$ which will satisfy the F-distribution with $k-1$ and $(k-1)(N-1)$ degrees of freedom can be written as:
\begin{equation}
	F_F = \frac{(N-1)\chi_F^2}{N(k-1)-\chi_F^2},
\end{equation}
in which
\begin{equation}
	\chi_F^2 = \frac{12N}{k(k+1)}\Bigg[\sum_{j=1}^{k}R_j^2-\frac{k(k+1)^2}{4}\Bigg].
\end{equation}
Table \ref{friedman} shows the Friedman statistics $F_F$ and the corresponding critical value in regard to each metric ($\#$ comparing algorithms $k=9$, $\#$ datasets $N=7$). With respect to each metric, the null hypothesis of equivalent performance among the compared methods can be rejected at the 0.05 significance level.

\begin{figure*}
	\centering
	\subfigure[ maF1] {\includegraphics[scale=0.5]{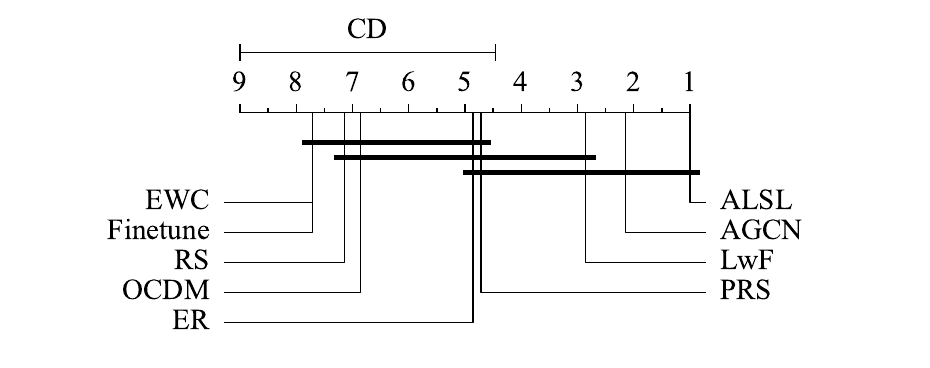}}
	\subfigure[ miF1] {\includegraphics[scale=0.5]{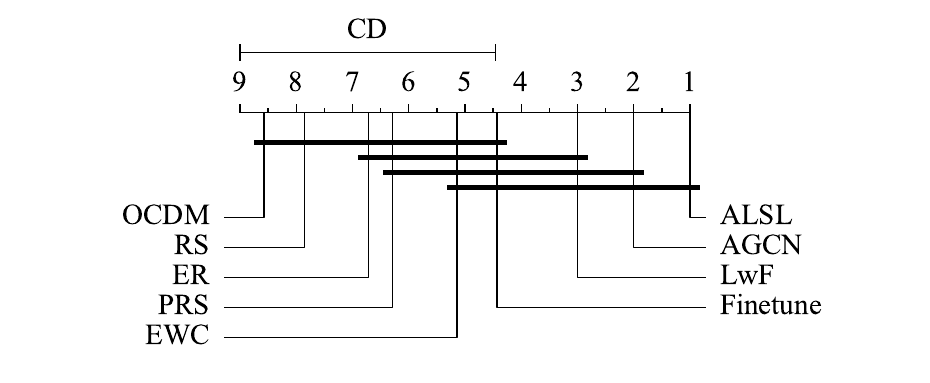}}
	\subfigure[ mAP] {\includegraphics[scale=0.5]{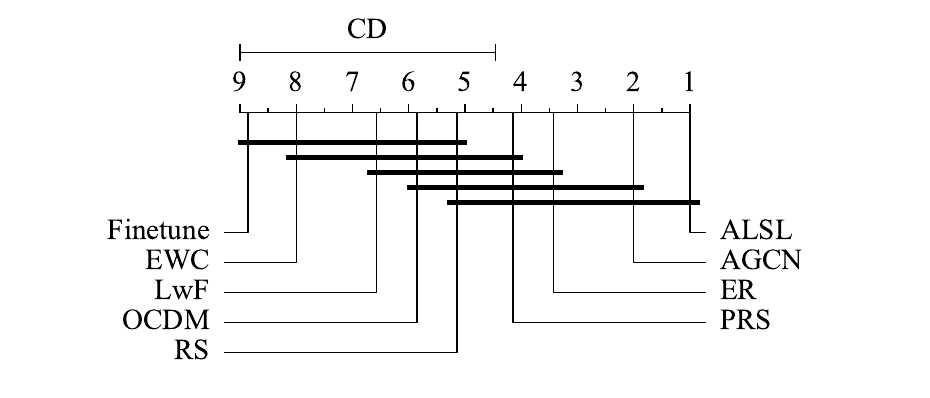}}
	\caption{Pairwise comparisons with the Nemenyi test in 7 datasets and 9 algorithms used in our experiments. Algorithms not connected with each other in the CD diagram are considered to have significantly different performance ($CD=4.540$ at 0.05 significance level).}
	\label{ne}
\end{figure*}

Then, we perform the strict post-hoc Nemenyi test \cite{demvsar2006statistical} which is used to account for pairwise comparisons for all compared approaches. The critical difference (CD) value of the rank difference between two algorithms is: 
\begin{equation}
	CD=q_{\alpha}\sqrt{\frac{k(k+1)}{6N}},
\end{equation}
in which $q_{\alpha}=3.102$ at 0.05 significance level. Therefore, one algorithm can be considered as having significantly different performance than another method if their average ranks difference is larger than CD ($CD=4.540$ in our experimental setting). Figure \ref{ne} reports the CD diagrams on each metric, where the average rank of each compared method is marked along the axis (the smaller the better). The algorithms that are not connected by a horizontal line are considered to have significant differences in performance. We can observe that: (1) In terms of mAP, AESL and other MLCIL approaches (except OCDM) significantly outperforms SLCIL methods. (2) In terms of miF1 and maF1, rehearsal-free methods are significantly better than rehearsal-based algorithms. (3) AESL is not significantly different from some MLCIL approaches. This is due to the factor that these approaches beats other comparing approaches and the Nemenyi test fails to detect that AESL achieves a consistently better average ranks than other methods on all evaluation metrics.

\section{Limitations}
At the application level, in the affective HCI tasks in real-life scenarios, in addition to learning new emotion categories, we also need to adapt to new subjects. Therefore, it is necessary to further consider the continuous learning of emotions from different subjects, which can be regarded as domain incremental learning. At the experimental level, the impact of the order of learning emotion categories and the number of emotion categories in each task on the experimental results needs to be further explored.

\section{Broader Impact}
Since our work does not involve AI ethics and does not involve private data, the broader impact is not available.

\begin{algorithm*}
	\caption{Training procedure of AESL.}
	\label{alg:AESL}
	\LinesNumbered
	\small\KwIn{Training sequence $\left\{\mathcal{D}^{1}, \cdots, \mathcal{D}^{B}\right\}$. Hyperparameters $\sigma$, $\beta$, $\lambda_1$, $\lambda_2$, $\lambda_3$. Affective dimension features $\left\lbrace \boldsymbol{\tau}^1, \cdots, \boldsymbol{\tau}^B \right\rbrace $.}
	\small
	\For{$b=1:B$}{
		\While{\textnormal{not converged}}{
			\For{$(\mathbf{x}^b,\mathbf{y}^b) \sim \mathcal{D}^b$}{
				\eIf{$b=1$}
				{Compute $\mathbf{A}^b$ directy with label matrix $\mathbf{Y}^b$ using Eq.\ref{label_co}.}
				{Compute soft label matrix $\mathbf{S}$ with $\mathbf{s} = \psi(\mathbf{x}^{b},\mathbf{A}^{b-1};\Phi^{b-1})$ and set intial label confidence matrix $\mathbf{F}_0=\mathbf{S}$.
					
					Compute the normalizing weight matrix $\hat{\mathbf{P}} = \mathbf{P}\mathbf{D}^{-1}$. $\mathbf{P}_{ij}$ is the similarity between two instances in $\mathcal{D}^b$.
					
					Implement label propagating to obatin refined soft label matrix $\hat{\mathbf{S}}$ with Eq.\ref{pro}.
					
					Compute $\mathbf{B}^b$ directy with label matrix $\mathbf{Y}^b$ using Eq.\ref{label_co}.
					
					Compute $\mathbf{R}^b$ and $\mathbf{Q}^b$ with $\hat{\mathbf{S}}$ and $\mathbf{Y}^b$ using Eq.\ref{rb} and \ref{qb}, then obtain ${\mathbf{A}}^{b}=\begin{bmatrix} {\mathbf{A}}^{b-1} & \mathbf{R}^{b} \\ \mathbf{Q}^{b} & \mathbf{B}^{b} \end{bmatrix}$.
					
					\tcp{\small Get augmented ERG shown in Section \ref{aERG}.}
				}
				
				Construct (augmented) ERG $\mathcal{G}^b$ with initial node features $\mathbf{H}^b_{0}$ and adjacency matrix $\mathbf{A}^{b}$.
				
				Implement message passing strategy using Eq.\ref{mp} to obtain label semantic embeddings $\mathbf{E}^b$.
				
				\vspace{0.1cm}
				\tcp{\small Implement label semantics learning shown in Section \ref{lsl}.}
				\vspace{0.1cm}
				
				Compute importance vectors $\mathbf{\alpha}$ using a fully-connected network followed by a $\mathrm{sigmoid}$ function.
				
				Compute semantic-specific features $\mathbf{o}$ with deep latent feature $\mathbf{z}$ and importance vectors $\mathbf{\alpha}$ with Eq.\ref{lsf}.
				
				Compute the label confidence scores $\mathbf{s}$ using Eq.\ref{confidence} to obtain the prediction for emotion classes $1,...,\mathcal{C}^b$.
				
				\vspace{0.1cm}
				\tcp{\small Implement semantic-guided feature decoupling to obtain semantic-specific features shown in Section \ref{sgfd}.}
				\vspace{0.1cm}
				
				Compute the representation similarity matrix $\mathbf{M}^b$, $\mathbf{M}^{b-1}$ and $\mathbf{M}^{\text{aff}}$ with Eq.\ref{rsmb} and \ref{rsmaff}.
				
				\vspace{0.1cm}
				\tcp{\small Implement relation-based knowledge distillation with affective dimension features shonw in Section \ref{rkd}.}
				\vspace{0.1cm}
				
				Compute the final loss $\mathcal{L}$ with Eq.\ref{le_loss}, \ref{rkd_loss} and \ref{final_loss}. Update AESL model by minimizing $\mathcal{L}$.
				
			}
		}
	}		
\end{algorithm*}

\begin{table*}[h]
	\renewcommand\arraystretch{1.1}
	\scriptsize
	\caption{Class incremental results on subject 2 of \emph{Brain27} dataset. AGCN, PRS, OCDM are MLCIL algorithm in these compared methods.}
	\resizebox{\textwidth}{!}{
		\begin{tabular}{l|c|ccc|c|ccc|c|ccc|c|ccc}
			\hline
			\multirow{3}{*}{\textbf{Method}}  & \multicolumn{4}{c|}{\textbf{Brain27 B0-I9}}  & \multicolumn{4}{c|}{\textbf{Brain27 B0-I3}} & \multicolumn{4}{c|}{\textbf{Brain27 B15-I3}}&\multicolumn{4}{c}{\textbf{Brain27 B15-I2}} \\ \cline{2-17} 
			&    \multicolumn{1}{c|}{Avg. Acc} & \multicolumn{3}{c|}{Last Acc}& \multicolumn{1}{c|}{Avg. Acc} & \multicolumn{3}{c|}{Last Acc} &\multicolumn{1}{c|}{Avg. Acc} & \multicolumn{3}{c|}{Last Acc}&\multicolumn{1}{c|}{Avg. Acc} & \multicolumn{3}{c}{Last Acc}\\ \cline{2-17}  
			&   mAP &  maF1 & miF1 & mAP    & mAP  &  maF1 & miF1 & mAP & mAP  &  maF1 & miF1 & mAP& mAP  &  maF1 & miF1 & mAP  \\   
			\midrule 
			Upper-bound  &    - &  34.0 & 44.3 & 42.3 &    -   &  34.0   &44.3 & 42.3 &  -& 34.0 & 44.3  & 42.3  & -  & 34.0  & 44.3  & 42.3 \\   
			\midrule
			Finetune  & 34.9  &8.2  &18.5  & 24.8 & 31.0 & 5.0 & 14.0 &21.0  & 25.8 & 5.1 & 14.4 &20.0  & 23.4 & 3.7 &13.3  &18.7  \\ 
			EWC   &  34.3 & 7.8 & 18.1 & 25.2 & 31.2 & 5.0 &13.7 & 21.7 & 26.4 & 5.0 & 13.8 & 20.1 & 24.1 & 3.7 & 13.3 & 19.4\\ 
			LwF   &  38.9 & 11.4 & 28.6 & 29.6 & 37.0 & 19.4 &37.0  & 25.9 & 31.8 & 15.7 & 32.5 & 24.4 & 29.4 & 15.4 & 32.6 &21.9 \\ 
			ER   &   40.4& 8.5 & 13.1 & 34.9 & 39.3 & 3.8 & 4.8 & 20.8 & 37.3 & 6.9 & 11.6 & 34.1 & 36.3 & 7.4 & 11.0 &  32.7 \\ 
			RS  &  41.0 & 7.6 & 10.8 &  35.9& 40.4 & 3.7 & 5.9 &  28.9& 36.4 & 6.2 & 10.3 & 31.4 & 35.8 & 4.3 & 7.8 & 31.0\\ 
			\midrule
			AGCN  &  43.4 & 28.8  & 43.1 & 39.8 & 43.0 & 32.4 & 43.6 &33.7  &39.0  &24.8  & 39.5 & 35.2 & 37.3 & 24.0 &35.5  & 32.1    \\ 
			PRS   & 40.4  & 7.9 & 14.4 & 34.4 & 40.9 & 3.6 & 7.8 & 20.2 & 37.8 & 8.4 & 12.0 & 34.1 & 37.2 & 8.5 & 11.8 & 32.5\\ 
			OCDM   &  40.0 & 8.4 & 12.9 & 33.4 & 40.5 & 4.5 & 8.1 & 28.2 & 36.1 & 5.9 & 9.2 & 30.9 & 34.7 & 5.1 & 8.6 &  29.4\\ 
			\textbf{AESL}   &\textbf{45.4}   &\textbf{29.8}  &\textbf{43.2}  & \textbf{41.4} & \textbf{45.5} &\textbf{32.5}  &\textbf{44.2}  &\textbf{35.3}  &\textbf{40.2}  & \textbf{25.5} & \textbf{40.1} &\textbf{36.4}  & \textbf{39.3}  & \textbf{24.8} & \textbf{36.4}  &  \textbf{35.0}      \\ 
			\hline
		\end{tabular}
	}
	\label{tb:brain27(2)}
\end{table*}
\begin{table*}[h]
	\renewcommand\arraystretch{1.1}
	\scriptsize
	\caption{Class incremental results on subject 3 of \emph{Brain27} dataset. AGCN, PRS, OCDM are MLCIL algorithm in these compared methods.}
	\resizebox{\textwidth}{!}{
		\begin{tabular}{l|c|ccc|c|ccc|c|ccc|c|ccc}
			\hline
			\multirow{3}{*}{\textbf{Method}}  & \multicolumn{4}{c|}{\textbf{Brain27 B0-I9}}  & \multicolumn{4}{c|}{\textbf{Brain27 B0-I3}} & \multicolumn{4}{c|}{\textbf{Brain27 B15-I3}}&\multicolumn{4}{c}{\textbf{Brain27 B15-I2}} \\ \cline{2-17} 
			&    \multicolumn{1}{c|}{Avg. Acc} & \multicolumn{3}{c|}{Last Acc}& \multicolumn{1}{c|}{Avg. Acc} & \multicolumn{3}{c|}{Last Acc} &\multicolumn{1}{c|}{Avg. Acc} & \multicolumn{3}{c|}{Last Acc}&\multicolumn{1}{c|}{Avg. Acc} & \multicolumn{3}{c}{Last Acc}\\ \cline{2-17}  
			&   mAP &  maF1 & miF1 & mAP    & mAP  &  maF1 & miF1 & mAP & mAP  &  maF1 & miF1 & mAP& mAP  &  maF1 & miF1 & mAP  \\   
			\midrule 
			Upper-bound  &    - & 33.6  & 44.2 & 42.0 &    -   &     33.6  & 44.2 & 42.0  &  -&    33.6  & 44.2 & 42.0   & -  &   33.6  & 44.2 & 42.0  \\   
			\midrule
			Finetune  &33.0   & 7.7 & 17.3 & 25.8 & 29.4 & 5.0 & 14.0 & 20.5 & 25.0 & 5.3 & 14.0 &20.0  & 23.2 & 4.1 & 13.8 &19.2  \\ 
			EWC       &32.4   & 7.7 & 16.7 & 25.3 & 29.6 & 5.2 & 13.8 & 19.8 & 25.5 & 5.3 & 14.0 & 20.0 & 24.0 & 4.0 & 13.6 &19.6 \\ 
			LwF       &36.4   & 11.7 & 28.7 & 30.3 & 34.7 & 21.7 & 38.3 & 23.5 & 30.3 & 15.4 & 31.5 & 23.6 & 28.2 & 15.6 & 31.8 &21.3 \\ 
			ER        & 37.1  & 7.0 & 10.2 & 33.5 & 36.7 & 2.4 & 2.4 & 30.5 & 34.6 & 5.0 & 8.6 & 31.9 & 35.4 & 6.8 & 10.9 & 32.9  \\ 
			RS        & 38.1  & 6.5 & 9.2 & 35.8 & 37.7 & 2.9 & 5.2 & 29.0 & 34.1 & 5.0 &7.7  & 30.3 & 33.8 & 3.6 & 6.8 & 30.0\\ 
			\midrule
			AGCN      & 40.2  &  27.6 & 42.0 & 38.2 & 40.1 & 32.0 & 42.9 & 32.7 & 37.0 & 27.0 & 37.1 & 33.6 & 34.9 & 24.1 & 36.6 &30.2     \\ 
			PRS       & 38.0  & 7.6 & 12.0 & 34.3 & 38.8 & 3.6 & 6.0 & 29.5 & 35.5 & 5.5 & 7.7 & 32.7 & 36.0 & 4.5 & 7.3 & 33.1\\ 
			OCDM      &  37.9 & 8.4 & 11.7 & 35.0 & 38.0 & 5.4 & 8.2 & 30.1 & 34.4 & 4.0 & 6.0 & 30.3 & 33.2 & 3.1 & 5.3 & 28.5 \\ 
			\textbf{AESL}   &\textbf{42.6}   &\textbf{28.5}  &\textbf{44.0}  & \textbf{40.8} & \textbf{42.3} & \textbf{33.4} &\textbf{43.2}  & \textbf{33.4} & \textbf{38.8}  & \textbf{27.2} &\textbf{38.9}  &\textbf{35.4} & \textbf{37.9} &  \textbf{25.5}& \textbf{37.7} & \textbf{34.1}      \\ 
			\hline
		\end{tabular}
	}
	\label{tb:brain27(3)}
\end{table*}

\begin{table*}[h]
	\renewcommand\arraystretch{1.1}
	\scriptsize
	\caption{Class incremental results on subject 4 of \emph{Brain27} dataset. AGCN, PRS, OCDM are MLCIL algorithm in these compared methods.}
	\resizebox{\textwidth}{!}{
		\begin{tabular}{l|c|ccc|c|ccc|c|ccc|c|ccc}
			\hline
			\multirow{3}{*}{\textbf{Method}}  & \multicolumn{4}{c|}{\textbf{Brain27 B0-I9}}  & \multicolumn{4}{c|}{\textbf{Brain27 B0-I3}} & \multicolumn{4}{c|}{\textbf{Brain27 B15-I3}}&\multicolumn{4}{c}{\textbf{Brain27 B15-I2}} \\ \cline{2-17} 
			&    \multicolumn{1}{c|}{Avg. Acc} & \multicolumn{3}{c|}{Last Acc}& \multicolumn{1}{c|}{Avg. Acc} & \multicolumn{3}{c|}{Last Acc} &\multicolumn{1}{c|}{Avg. Acc} & \multicolumn{3}{c|}{Last Acc}&\multicolumn{1}{c|}{Avg. Acc} & \multicolumn{3}{c}{Last Acc}\\ \cline{2-17}  
			&   mAP &  maF1 & miF1 & mAP    & mAP  &  maF1 & miF1 & mAP & mAP  &  maF1 & miF1 & mAP& mAP  &  maF1 & miF1 & mAP  \\   
			\midrule 
			Upper-bound  &    - & 38.3  & 48.6 &45.1  &    -   &    38.3  & 48.6 &45.1  &  -&   38.3  & 48.6 &45.1   & -  &   38.3  & 48.6 &45.1 \\   
			\midrule
			Finetune  & 36.0  & 7.9 & 19.0 & 25.2 &30.9  & 4.9 & 13.7 & 21.1 &26.4  & 4.6 & 13.4 & 20.3 & 23.9 & 3.7 & 13.2 &18.1  \\ 
			EWC     &  35.0 & 7.9 & 18.5 & 25.3 & 31.7 & 5.1 & 13.8 & 22.0 & 27.2 & 4.8 &13.7  & 20.4 & 25.1 & 3.7 & 13.2 & 20.1\\ 
			LwF     &  39.6 & 15.1 & 33.9 & 30.8 & 37.5 & 20.5 & 37.5 & 23.1 & 33.3 &17.2  & 32.8 & 25.2 & 30.9 & 15.8 & 33.5 & 22.7\\ 
			ER     & 42.0  & 10.1 & 15.6 & 37.7 & 41.8 & 3.9 & 4.8 & 34.8 & 40.1 & 9.3 & 11.3 & 37.0 & 38.8 & 9.7 & 12.4 &  36.4 \\ 
			RS    &  42.6 & 9.0 & 14.2 & 38.1 & 43.5 & 4.3 & 7.1 & 34.4 & 38.8 & 7.5 & 10.8 & 34.6 & 37.8 & 5.4 & 10.6 & 33.2\\ 
			\midrule
			AGCN    & 43.7  & 31.8  & 46.4 & 40.9 & 44.1 & 36.1 & 44.5 & 36.3 & 41.0 & 30.3 & 43.7 & 38.6 & 39.3 & 26.6 & 37.1 &35.0     \\ 
			PRS     &  43.5 & 10.7 & 15.9 & 39.9 & 44.2 & 4.0 & 5.6 & 32.8 &40.2  & 8.5 & 11.1 & 36.1 & 39.3 & 9.1 & 12.2 & 35.6\\ 
			OCDM    &  43.4 & 11.4 & 15.1 & 39.8 & 43.0 & 4.9 & 7.8 & 31.7 & 38.2 & 6.0 & 5.9 & 33.4 & 38.4 & 5.2 & 9.4 & 33.4 \\ 
			\textbf{AESL}   & \textbf{45.7}  & \textbf{36.4} &\textbf{47.3}  &\textbf{44.0}  & \textbf{47.1} & \textbf{37.2} & \textbf{46.4} & \textbf{38.7}  &\textbf{43.4} &\textbf{31.3}  & \textbf{44.4} & \textbf{39.9} &\textbf{42.2}  &  \textbf{29.0}&\textbf{40.4}  &  \textbf{37.7}     \\ 
			\hline
		\end{tabular}
	}
	\label{tb:brain27(4)}
\end{table*}

\begin{table*}[h]
	\renewcommand\arraystretch{1.1}
	\scriptsize
	\caption{Class incremental results on subject 5 of \emph{Brain27} dataset. AGCN, PRS, OCDM are MLCIL algorithm in these compared methods.}
	\resizebox{\textwidth}{!}{
		\begin{tabular}{l|c|ccc|c|ccc|c|ccc|c|ccc}
			\hline
			\multirow{3}{*}{\textbf{Method}}  & \multicolumn{4}{c|}{\textbf{Brain27 B0-I9}}  & \multicolumn{4}{c|}{\textbf{Brain27 B0-I3}} & \multicolumn{4}{c|}{\textbf{Brain27 B15-I3}}&\multicolumn{4}{c}{\textbf{Brain27 B15-I2}} \\ \cline{2-17} 
			&    \multicolumn{1}{c|}{Avg. Acc} & \multicolumn{3}{c|}{Last Acc}& \multicolumn{1}{c|}{Avg. Acc} & \multicolumn{3}{c|}{Last Acc} &\multicolumn{1}{c|}{Avg. Acc} & \multicolumn{3}{c|}{Last Acc}&\multicolumn{1}{c|}{Avg. Acc} & \multicolumn{3}{c}{Last Acc}\\ \cline{2-17}  
			&   mAP &  maF1 & miF1 & mAP    & mAP  &  maF1 & miF1 & mAP & mAP  &  maF1 & miF1 & mAP& mAP  &  maF1 & miF1 & mAP  \\   
			\midrule 
			Upper-bound  &    - &  36.6 & 46.8 & 45.0 &    -   &    36.6 & 46.8 & 45.0  &  -&    36.6 & 46.8 & 45.0   & -  &  36.6 & 46.8 & 45.0 \\   
			\midrule
			Finetune  & 34.7  &7.3  & 17.8 &25.4  & 31.1 & 5.4 & 14.3 & 21.7 & 25.4 & 5.2 & 13.9 & 18.9 & 23.4 &3.9  &13.2  & 19.2 \\ 
			EWC      & 34.1  & 6.7 & 15.6 & 25.1 & 31.0 & 4.9 & 13.5 & 21.9 & 26.3 & 5.0 & 13.7 & 20.3 & 24.3 & 4.0 & 13.5 & 20.2\\ 
			LwF      &  37.6 & 12.2 & 30.3 & 28.7 & 35.5 & 21.0 & 37.9 & 25.8 & 31.2 & 15.8 & 31.7 & 24.6 & 29.4 & 14.3 & 29.0 & 22.4\\ 
			ER      &  40.7 & 8.8 & 13.9 & 35.8 & 41.7 & 4.5 & 7.6 & 34.0 & 38.1 & 6.9 & 8.9 & 36.4 & 36.3 & 6.9 & 9.8 & 32.9  \\ 
			RS     &  41.4 & 7.6 & 11.3 & 36.9 & 41.5 & 4.4 & 7.6 & 31.7 & 37.1 & 5.7 & 8.6 & 33.4 & 37.3 & 5.4 & 9.0 & 33.0\\ 
			\midrule
			AGCN    & 43.8  &  29.5 & 44.0 & 41.4 & 42.9 & 32.5 & 42.5 & 34.1 & 38.4 & 28.7 & 41.5 & 35.0 & 37.3 & 24.6 & 37.6 &33.1     \\ 
			PRS     &  41.2 & 8.9 & 16.4 & 37.0 & 42.4 & 5.3 & 8.3 & 32.0 & 37.8 & 6.7 & 9.0 & 34.1 & 38.4 & 6.5 & 9.4 & 34.9\\ 
			OCDM    &  41.1 & 8.3 & 14.0 & 35.5 & 41.3 & 5.4 & 8.0 & 29.7 & 36.8 & 4.4 & 7.9 & 31.5 & 34.8 & 4.6 & 8.0 &  30.1\\ 
			\textbf{AESL}   & \textbf{46.0}  & \textbf{32.9} & \textbf{45.8} &\textbf{44.1}  &\textbf{45.5}  &\textbf{35.3} &\textbf{45.6}  &\textbf{35.4}  &\textbf{40.4}  &\textbf{29.5}  & \textbf{44.5} & \textbf{37.5} &\textbf{39.9}  &  \textbf{29.3}& \textbf{40.8} & \textbf{36.2}  \\ 
			\hline
		\end{tabular}
	}
	\label{tb:brain27(5)}
\end{table*}

\end{document}